\documentclass{article}

    \usepackage[preprint]{neurips_2025}

\usepackage{booktabs}
\usepackage[table]{xcolor}
\usepackage{tabularx}  
\usepackage{threeparttable}  
\usepackage{booktabs} 
\usepackage{multirow} 
\usepackage{array}   
\usepackage{placeins}
\usepackage{xcolor}
\usepackage{enumitem}
\usepackage{amsmath} 
\definecolor{golden}{rgb}{1.0, 0.843, 0.0}
\definecolor{lightblue}{rgb}{0.75, 0.75, 0.75}
\definecolor{lightgreen}{rgb}{0.8, 0.5, 0.2}

\usepackage{times}
\usepackage{latexsym}
\usepackage{graphicx}
\usepackage{pifont}
\usepackage{makecell}
\usepackage{amsmath}   
\usepackage{amssymb}   
\usepackage{bm}        

\usepackage{authblk}
\usepackage{microtype}
\usepackage{inconsolata}

\usepackage[utf8]{inputenc} 
\usepackage[T1]{fontenc}    
\usepackage{hyperref}       
\usepackage{url}            
\usepackage{booktabs}       
\usepackage{amsfonts}       
\usepackage{nicefrac}       
\usepackage{microtype}      
\usepackage{xcolor}         

\newcommand{\TheName}{BizFinBench}
\newcommand{\TheEvaName}{IteraJudge}
\newcommand{\TitleLogo}{%
  \raisebox{-0.1\height}{\includegraphics[height=0.8cm]{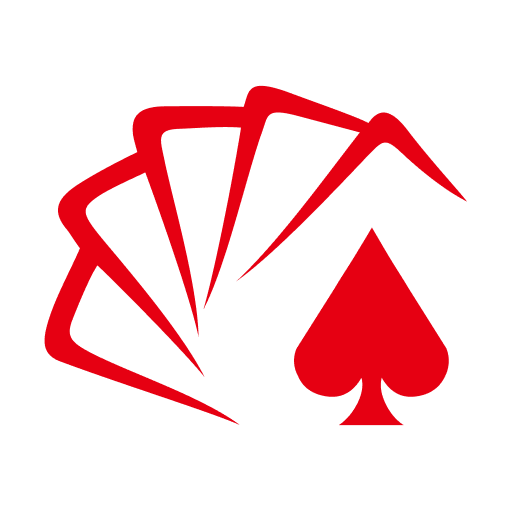}}%
}

\title{
  \TitleLogo\hspace{0.05em}\TheName{}: A Business-Driven Real-World Financial Benchmark for Evaluating LLMs
}

\author[1]{Guilong Lu$^*$}
\author[1,2]{Xuntao Guo$^*$}
\author[1]{Rongjunchen Zhang$^{\spadesuit}$}
\author[1]{Wenqiao Zhu}
\author[1]{Ji Liu}

\affil[1]{HiThink Research}
\affil[2]{Harbin Institute of Technology}

\begin{document}

\maketitle
\def\customfootnotetext#1#2{{%
  \let\thefootnote\relax
  \footnotetext[#1]{#2}}}
\customfootnotetext{1}{$^*$ denotes equal contribution. This paper was completed during Guilong Lu and Xuntao Guo's internships at Hithink Research. }
\customfootnotetext{1}{$\spadesuit$ denotes corresponding author, zhangrongjunchen@myhexin.com.}

\begin{abstract}
\label{abstract}
Large language models excel in general tasks, yet assessing their reliability in logic‑heavy, precision‑critical domains like finance, law, and healthcare remains challenging. To address this, we introduce \TheName{}, the first benchmark specifically designed to evaluate LLMs in real-world financial applications. \TheName{} consists of 6,781 well-annotated queries in Chinese, spanning five dimensions: numerical calculation, reasoning, information extraction, prediction recognition, and knowledge-based question answering, grouped into nine fine-grained categories. The benchmark includes both objective and subjective metrics. We also introduce \TheEvaName{}, a novel LLM evaluation method that reduces bias when LLMs serve as evaluators in objective metrics. We benchmark 25 models, including both proprietary and open-source systems. Extensive experiments show that no model dominates across all tasks. Our evaluation reveals distinct capability patterns: (1) In Numerical Calculation, Claude-3.5-Sonnet (63.18) and DeepSeek-R1 (64.04) lead, while smaller models like Qwen2.5-VL-3B (15.92) lag significantly; (2) In Reasoning, proprietary models dominate (ChatGPT-o3: 83.58, Gemini-2.0-Flash: 81.15), with open-source models trailing by up to 19.49 points; (3) In Information Extraction, the performance spread is the largest, with DeepSeek-R1 scoring 71.46, while Qwen3-1.7B scores 11.23; (4) In Prediction Recognition, performance variance is minimal, with top models scoring between 39.16 and 50.00. We find that while current LLMs handle routine finance queries competently, they struggle with complex scenarios requiring cross-concept reasoning. BizFinBench offers a rigorous, business-aligned benchmark for future research. The code and dataset are available at https://github.com/HiThink-Research/BizFinBench.

\end{abstract}

\section{Introduction}
\label{Introduction}

Recent years have witnessed rapid advancements of Large Language Models (LLMs), which demonstrates remarkable capabilities across diverse domains, such as finance, law, healthcare and so on~\cite{chen2024survey, hithinkliu2025nexus, hithinkzhang2023dynalogue, lu2024grace, xie2025medical}. In financial applications, LLMs are increasingly applied to complex tasks, including automated financial analysis, fraud detection, risk assessment, and investment strategy formulation~\cite{zhao2024revolutionizing, gan2024mme}. However, evaluating the robustness and reliability of LLMs in finance domains remains a significant challenge.

\begin{figure*}
    \centering
    \includegraphics[scale=0.06]{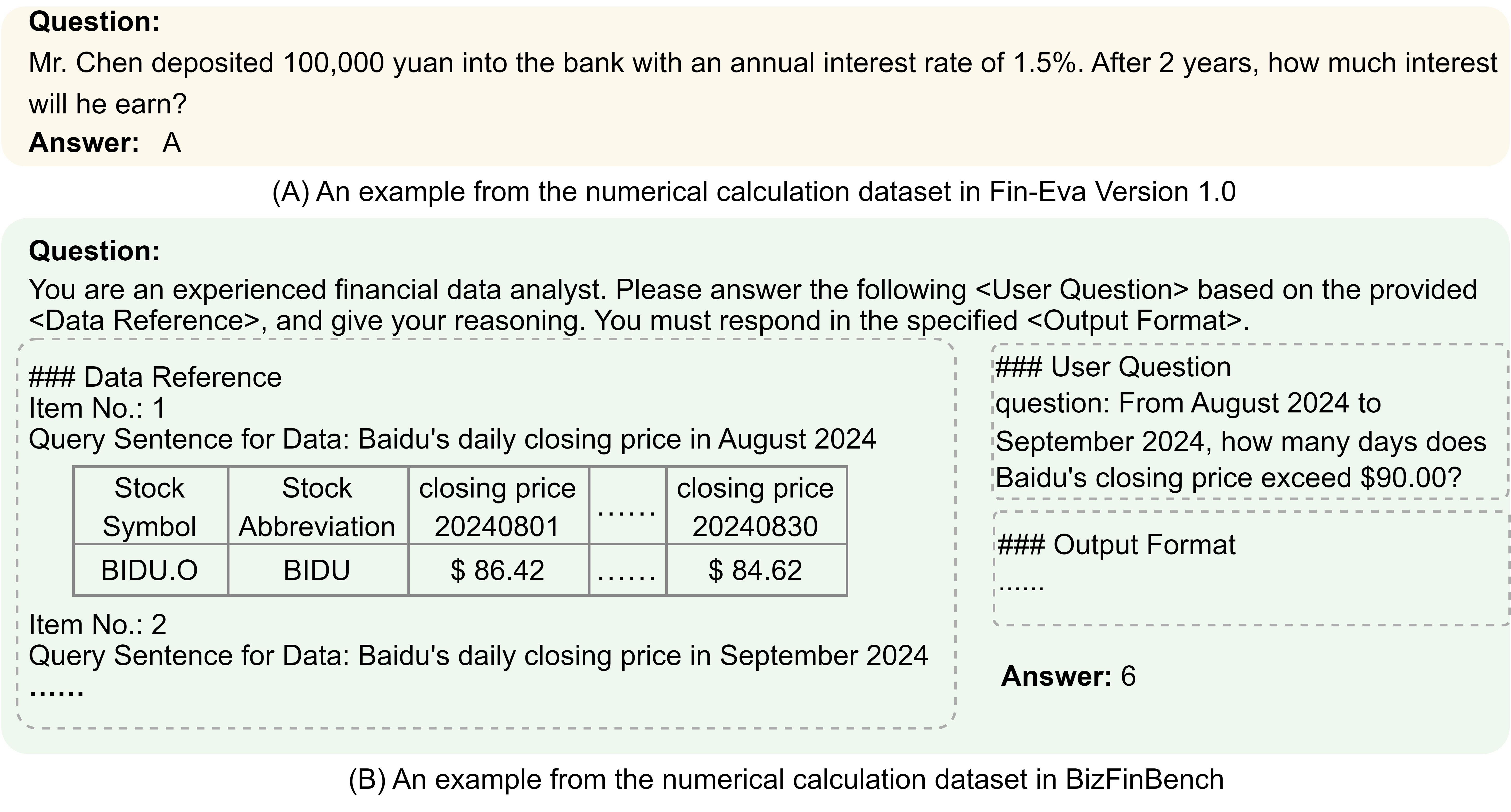}
    \caption{Comparison of numerical calculation questions in Fin-Eva~\cite{team2023FinEva} and \TheName{}. The Fin-Eva example presents a straightforward financial math problem, while the \TheName{} example requires multi-step reasoning: first analyzing the problem, then extracting and utilizing relevant data from a provided markdown-formatted table for accurate computation. An Chinese version is included in the Appendix for clarity and ease of reference.}
    \label{fig:motivationenglish}
\end{figure*}

Different from traditional Science, Technology, Engineering, and Mathematics (STEM) questions, where inputs are typically short, well-structured, and yield deterministic answers, financial tasks are more complex. They typically involve long context, structured inputs (e.g., tabular stock data, market news), require temporal reasoning, and demand fine-grained judgment under ambiguity. As illustrated in Figure~\ref{fig:motivationenglish}, STEM-style questions usually have clear computational logic and a single correct answer, while financial tasks call for multi-step reasoning over real-world data, generally with adversarial or noisy context~\cite{du2024evaluation}.

Despite the emergence of financial benchmarks such as FinEval~\cite{FinEval}, existing approaches treat financial tasks as general document Query-Answering (QA)~\cite{wang2024leave}, lacking structured inputs and business-grounded reasoning required in practice. Thus, there emerges the gap between benchmark performance and real-world applicability.

To address these limitations, we introduce \TheName{}, a comprehensive benchmark designed to rigorously evaluate LLMs across a broad spectrum of real-world financial tasks. In contrast to previous benchmarks, \TheName{} adopts a business-driven data construction methodology and emphasizes contextual complexity and adversarial robustness. It encompasses five key dimensions: QA, prediction \& recognition, reasoning, information extraction, and numerical calculation. Under these dimensions, \TheName{} comprises nine distinct categories: anomalous event attribution, financial numerical computation, financial time reasoning, financial tool usage, financial knowledge QA, financial data description, emotional value evaluation, stock price prediction, and financial named entity recognition.

A core characteristic of \TheName{} is the focus on business-contextual evaluation. For example, in the anomalous event attribution task, LLMs are required to identify the causes of stock price anomalies by analyzing time-sensitive news feeds, some of which are deliberately embedded with misleading positive or negative information. This setting challenges LLMs to perform fine-grained reasoning and signal discrimination under realistic noise and uncertainty.

In addition to the benchmark design, a critical component of \TheName{} is the design of a reliable evaluation methodology. While constructing realistic tasks is essential, evaluating LLM outputs, particularly for open-ended, complex financial problems, remains a significant challenge.

\begin{table*}[htbp]
  \centering
  \caption{Comparison Between \TheName{} and Other Financial Datasets}
  \resizebox{\textwidth}{!}{%
    \begin{tabular}{lrp{13.835em}rllr}
    \toprule
    Data  & \multicolumn{1}{l}{Year} & \multicolumn{1}{l}{Task} & \multicolumn{1}{l}{Examples} & Language & Source & \multicolumn{1}{l}{Business-based} \\
    \midrule
    FLUE  & 2022  & Multiple financial NLP tasks & 26292 & English & Aggregated from existing sources & \ding{55} \\
    FLARE & 2023  & Multiple financial NLP tasks, \newline{}financial prediction tasks & 19196 & Chinese, English & Aggregated from existing sources & \ding{55} \\
    CF-Benchmark & 2024  & \multicolumn{1}{l}{Multiple financial NLP tasks} & 3917  & Chinese & except & \ding{55} \\
    FinEval & 2023  & \multicolumn{1}{l}{Multiple financial NLP tasks} & 8351  & Chinese & Financial field examination \& except & \ding{55} \\
    FinQA & 2021  & \multicolumn{1}{l}{Financial numerical reasoning} & 8281 & English & except & \ding{55} \\
    FinancelQ & 2023  & \multicolumn{1}{l}{Multiple financial NLP tasks} & 7137  & Chinese & Financial field examination \& except & \ding{55} \\
    CGCE  & 2023  & \multicolumn{1}{l}{Multiple financial NLP tasks} & 150   & Chinese, English & except & \ding{55} \\
    \TheName{} & 2025  & Multiple financial NLP tasks, \newline{}financial prediction tasks & 7016  & Chinese & except & \ding{51} \\
    \bottomrule
    \end{tabular}%
    }
  \label{datasetintro}%
\end{table*}%

Traditional human evaluation provides high-quality judgments but suffers from two major drawbacks: (1) the annotation cost increases exponentially with the scale and domain specificity of financial tasks, and (2) subjective inconsistencies among annotators can introduce substantial noise. Although recent approaches like LLM-as-a-Judge~\cite{gu2024survey} attempt to automate evaluation through prompt-based simulations of human judgment, they are prone to prompt bias and generally lack alignment with expert-level assessments. These limitations are further magnified in the financial domain, where tasks demand multi-step reasoning, contextual interpretation, and robustness against adversarial or misleading signals. As such, existing evaluation paradigms are insufficient to capture the depth and nuance required for trustworthy assessment.


To address this gap, we propose \textit{\TheEvaName{}}, an iterative calibration-based evaluation framework tailored for financial LLM benchmarks. Drawing inspiration from the RevisEval framework~\cite{zhang2024reviseval}, \textit{\TheEvaName{}} enhances evaluation accuracy and reliability through three core mechanisms: evaluation dimension disentanglement, sequential correction generation, and reference-aligned assessment. By integrating \textit{\TheEvaName{}} into \TheName{}, we establish a rigorous and interpretable evaluation pipeline for LLM performance in high-stakes financial contexts.

In summary, the major contributions of our work are as follows: 
\begin{itemize}
    \item We propose \TheName{}, the first evaluation benchmark in the financial domain that integrates business-oriented tasks, covering 5 dimensions and 9 categories. It is designed to assess the capacity of LLMs in real-world financial scenarios.
    \item We design a novel evaluation method, i.e., \TheEvaName{}, which enhances the capability of LLMs as a judge by refining their decision boundaries in specific financial evaluation tasks.
    \item We conduct a comprehensive evaluation with 25 LLMs based on \TheName{}, uncovering key insights into their strengths and limitations in financial applications.
\end{itemize}

\section{Related Work}
\label{related work}

In this section, we present existing evaluation benchmarks in financial domains. Then, we present the major LLMs specialized in financial domains.

\subsection{Financial Evaluation Benchmarks}

FLUE~\cite{shah2022flue} is a comprehensive suite of benchmarks covering five key financial tasks: sentiment analysis, news headline classification, named entity recognition, structural boundary detection, and question answering. Building on FLUE, FLARE~\cite{xie2023pixiu} expands the evaluation to include time-series processing capabilities, adding tasks such as stock price movement prediction.

In addition to FLUE and FLARE, several specialized datasets focus on various aspects of financial evaluation. For example, FinQA~\cite{chen2022finqa} provides QA pairs annotated by financial experts, accompanied by earnings reports from S\&P 500 companies. This dataset supports financial question answering, emphasizing detailed, factual responses based on corporate financial data. ConvFinQA~\cite{chen2022convfinqa} extends this by incorporating multi-turn dialogues, enabling more sophisticated interactions within the context of earnings reports, thus broadening the scope of financial evaluation to conversational contexts.


FinEval~\cite{FinEval} adopts a quantitative evaluation approach, combining long-term research insights with manual curation and featuring diverse question types. However, it primarily emphasizes static knowledge assessment and lacks coverage of dynamic, real-time financial tasks and fine-grained capability diagnostics, which limits its effectiveness in benchmarking models under complex, business-driven financial scenarios.

Expanding beyond traditional financial instruments such as stocks, bonds, and mutual funds, FinancelQ~\cite{financeIQ2023} introduces emerging topics such as cryptocurrencies and blockchain technologies. This dataset can be exploited for evaluating models in the rapidly evolving field of digital finance.

\begin{figure*}
    \centering
    \includegraphics[width=0.9\linewidth]{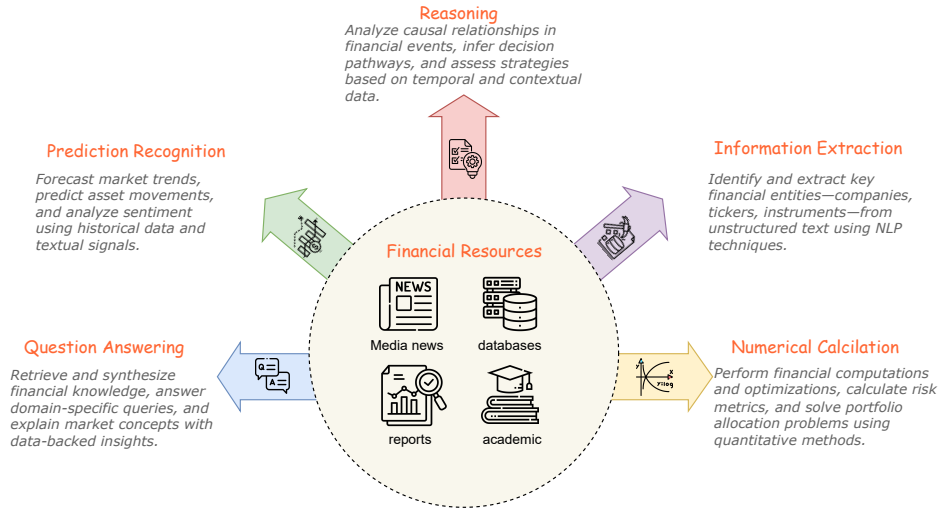}
    \caption{Distribution of tasks in \TheName{} across five key dimensions. The benchmark is structured around five dimensions, each focusing on a distinct capability of financial large language models. The figure also briefly illustrates the core focus of each dimension.}
    \label{distribution}
\end{figure*}

In the context of Chinese financial benchmarks, several recent datasets have been released, including CFBenchmark~\cite{lei2023cfbenchmark}, which focuses on Chinese financial text analysis; DISC-FINSFT~\cite{chen2023disc}, designed for financial sentiment analysis and forecasting; and CGCE~\cite{zhang2023cgce}, which extends financial evaluation to include general knowledge and commonsense reasoning in Chinese financial documents.

Table~\ref{datasetintro} provides a comprehensive comparison of existing financial benchmarks, detailing key aspects such as the year of release, the number of samples, language coverage, data sources, and whether the dataset was constructed with real business scenarios in mind. From the comparison, it is evident that while several benchmarks focus on financial knowledge or specific task types, they often rely on synthetic data or public information without a strong connection to actual business applications. In contrast, \TheName{} is the only benchmark explicitly designed around real-world financial operations and user interactions, making it uniquely positioned to evaluate the practical effectiveness of LLMs in authentic business environments. This business-centric design ensures higher relevance, realism, and applicability of the tasks included in the benchmark.

\subsection{Financial Large Language Models}

By training on a large corpus of financial data based on BERT, FinBERT~\cite{araci2019finbert} was proposed as a pre-trained model for the financial domain, primarily used for sentiment analysis of financial texts. Subsequently, models such as FinMA~\cite{xie2023pixiu}, InvestLM~\cite{yang2023investlm}, and FinGPT~\cite{yang2023fingpt} were fine-tuned on LLaMA~\cite{touvron2023llama1} to further enhance their performance in the financial domain. The XuanYuan3-70B model, built on the LLaMA3-70B architecture and incrementally pre-trained with a vast amount of Chinese and English corpora, focuses on the financial sector and is capable of handling complex tasks such as financial event interpretation, investment research applications, compliance, and risk management.
BloombergGPT~\cite{wu2023bloomberggpt} is a 50-billion-parameter LLM based on the Bloom architecture, specifically designed for the financial industry, demonstrating strong adaptability in the financial domain. Meanwhile, Baichuan4-Finance~\cite{zhang2024baichuan4} has achieved an accuracy rate of over 95\% in various certification fields such as banking, insurance, funds, and securities, further proving its exceptional performance in the vertical financial sector. Dianjin-R1~\cite{zhu2025dianjin} is designed for complex financial reasoning tasks and incorporates structured supervision along with dual-reward reinforcement learning, enabling it to outperform strong baselines across a range of financial benchmarks.

In addition, we also consider general-purpose models in our experiments, as many of them have undergone pre-training on datasets that contain financial texts, such as GPT-4o.


\begin{figure*}
    \centering
    \includegraphics[width=0.95\linewidth]{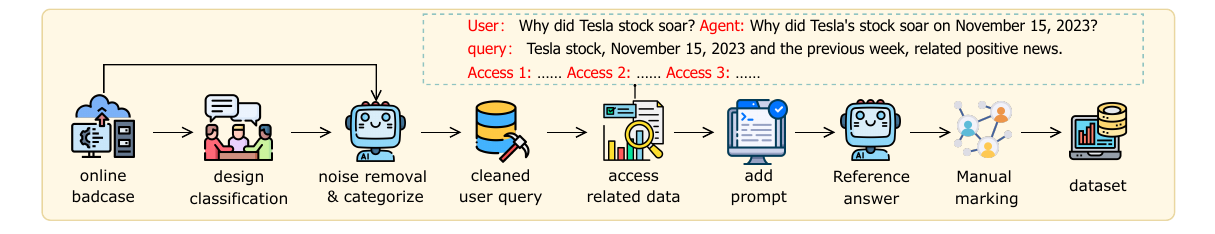}
    \caption{Workflow of \TheName{} dataset construction.}
    \label{datasetpip}
\end{figure*}

\section{\TheName{}}


In this section, we detail the design of \TheName{}, a comprehensive benchmark specialized for evaluating LLMs in financial domains. Compared to previous datasets, \TheName{} places a strong emphasis on business practicality and real-world applicability, aiming to bridge the gap between academic evaluation and the complex challenges encountered in real-world financial scenarios.

To capture the multifaceted nature of financial intelligence, we organize the benchmark into 10 distinct task types, which are further grouped into 5 overarching evaluation dimensions. As illustrated in Figure~\ref{distribution}~\footnote{An English version is included in the Appendix~\ref{other}}, these dimensions reflect key capabilities required in financial applications. For instance, the numerical computation dimension includes tasks that require models to perform financial computations and optimizations, calculate risk metrics, and solve portfolio allocation problems using quantitative methods. This dimension is designed to evaluate the capability of LLMs to apply precise mathematical reasoning in realistic financial contexts, where accuracy and analytical rigour are critical.
This structured categorization not only facilitates a fine-grained assessment of model strengths and weaknesses but also ensures that each component of the benchmark aligns with practical demands observed in financial services and business analytics.

\subsection{Data Construction}
\label{Data Construction}
Our dataset is primarily sourced from real user queries on the iwencai APP \footnote{https://www.iwencai.com/}, the APP serves a broad user base of individual investors and financial professionals, offering functionalities such as stock screening, market analysis, and personalized investment assistance. Leveraging advanced Artificial Intelligence (AI) technologies, iwencai enables users to perform complex financial analyses through natural language queries, covering areas like A-shares, Hong Kong and U.S. stocks, ETFs, and macroeconomic indicators.

Based on an extensive analysis of user queries from Platform A, our financial experts identified nine representative task categories that frequently appear in real-world financial scenarios. These include time reasoning, numerical computation, sentiment analysis, and so on. Notably, these categories collectively account for over 90\% of the queries observed on the platform, making them highly representative of actual business needs in financial decision-making.

To construct our dataset, we first aggregate a large set of real user queries, then employ GPT-4o~\cite{openai2023gpt4} to clean noisy entries, filter out incomplete or invalid ones, and classify each valid query into the appropriate expert-defined category. For underrepresented categories, we further use GPT-4o to synthesize additional data, ensuring category balance and coverage. This process results in a high-quality dataset tailored to practical financial applications.

\begin{table}[h]
  \centering
  \caption{Overview of \TheName{} Datasets}
  \resizebox{\textwidth}{!}{%
    \begin{tabular}{cp{17.8em}p{14.445em}p{5.39em}ll}
    \toprule
    Category & \multicolumn{1}{l}{Data} & \multicolumn{1}{l}{Evaluation Dimensions} & \multicolumn{1}{l}{Metrics} & Numbers & Avg Len. \\
    \midrule
    Reasoning & Anomalous Event Attribution (AEA) & Causal consistency\newline{}Information relevance\newline{}Noise resistance & Accuracy & 1064  & 939 \\
          & Financial Time Reasoning (FTR) & Temporal reasoning correctness & Accuracy & 514   & 1162 \\
          & Financial Tool Usage (FTU) & Tool selection appropriateness\newline{}Parameter input accuracy\newline{}Multi-tool coordination & Judge Score & 641   & 4556 \\
    \midrule
    Numerical calculation & Financial Numerical Computation (FNC) & Computational accuracy\newline{}Unit consistency & Accuracy & 581   & 651 \\
    \midrule
    Q\&A  & Financial Knowledge QA (FQA) & Question comprehension\newline{}Knowledge coverage\newline{}Answer accuracy & Judge Score & 990   & 22 \\
          & Financial Data Description (FDD) & Trend accuracy\newline{}Data consistency & Judge Score & 1461  & 311 \\
    \midrule
    Prediction recognition & Emotion Recognition (ER) & Emotion classification accuracy\newline{}Implicit information extraction & Accuracy & 600   & 2179 \\
          & Stock Price Prediction (SP) & Trend judgment, Causal reasoning & Accuracy & 497   & 4498 \\
    \midrule
    Information extraction & Financial Named Entity Recognition (FNER) & Recognition accuracy\newline{}Entity classification correctness & Accuracy & 435   & 533 \\
    \bottomrule
    \end{tabular}%
    }
  \label{datasetall}%
\end{table}%

Next, we collect relevant contextual data from internal financial databases and external sources based on the content of each user query. This includes stock prices, historical trading data, financial news, company disclosures, and so on. These sources are directly related to the query topic. For example, as illustrated in Figure~\ref{datasetpip}, when a user poses the question "Why did Tesla stock soar?", the raw query typically lacks explicit temporal markers. To address this problem, we automatically retrieve the timestamp, at which the query was originally issued and utilize it to construct a temporally anchored version of the query. This allows us to retrieve the most relevant financial information and news surrounding that specific point in time.

To enhance the discriminative power of the dataset, we carefully introduce designed distractor data into the context. These distractors are chosen to assess the reasoning capabilities of the model and include misleading but plausible information, such as news articles from unrelated companies, articles with opposing market sentiment (such as negative news during a stock rally), or temporally misaligned events. This step ensures that answering the question correctly requires understanding both the financial context and the time-sensitive nature of the data, rather than relying on superficial keyword matches.

Once the query and context are constructed, they are paired with a task-specific prompt and submitted to a large language model, such as GPT-4o, to generate candidate answers. These preliminary answers are not directly included in the dataset. Instead, each data point undergoes a rigorous human annotation and validation process.

To ensure a high level of data quality and reliability, every entry is independently reviewed and annotated by three senior financial experts. Each expert has over five years of professional experience in roles such as equity research, investment analysis, or portfolio management, and has previously worked at top-tier financial institutions, including securities firms, asset management companies, or banks. During annotation, the experts assess the accuracy of the model-generated answers and verify whether the task category assigned to the query is appropriate.

A data point is accepted into the final dataset only when all three experts reach full consensus across all aspects, including answer validity, contextual consistency, and category correctness. If disagreements arise, the entry is subjected to further review and iterative refinement until a unanimous agreement is reached. This multi-layered annotation process ensures that the dataset is not only factually accurate but also aligned with real-world financial reasoning and application standards.

\subsection{Statistics}
\label{sec:Dataset Statistics}

The \TheName{} benchmark consists of a total of 6,781 entries, encompassing a wide variety of tasks designed to assess model performance across diverse financial challenges. By testing models on these tasks, we aim to evaluate not only their individual capabilities but also their ability to generalize across multiple facets of financial data analysis.

Table~\ref{datasetall} provides a detailed breakdown of the dataset, including the evaluation dimensions, corresponding metrics, the number of instances per task, and the average token length per entry~\footnote{Detailed dataset information is provided in Appendix~\ref{appDataset details}.}. The dataset exhibits significant variability in input length, ranging from just 22 tokens to as many as 4,556 tokens. This broad range reflects the complexity and heterogeneity of real-world financial scenarios and presents a meaningful challenge for models to demonstrate their ability to process both short and long financial texts effectively.


\subsection{\TheEvaName{}: An Incremental Multi-Dimensional Evaluation Framework}
\label{sec:IteraJudge}

\begin{figure}
    \centering
    \includegraphics[width=1\linewidth]{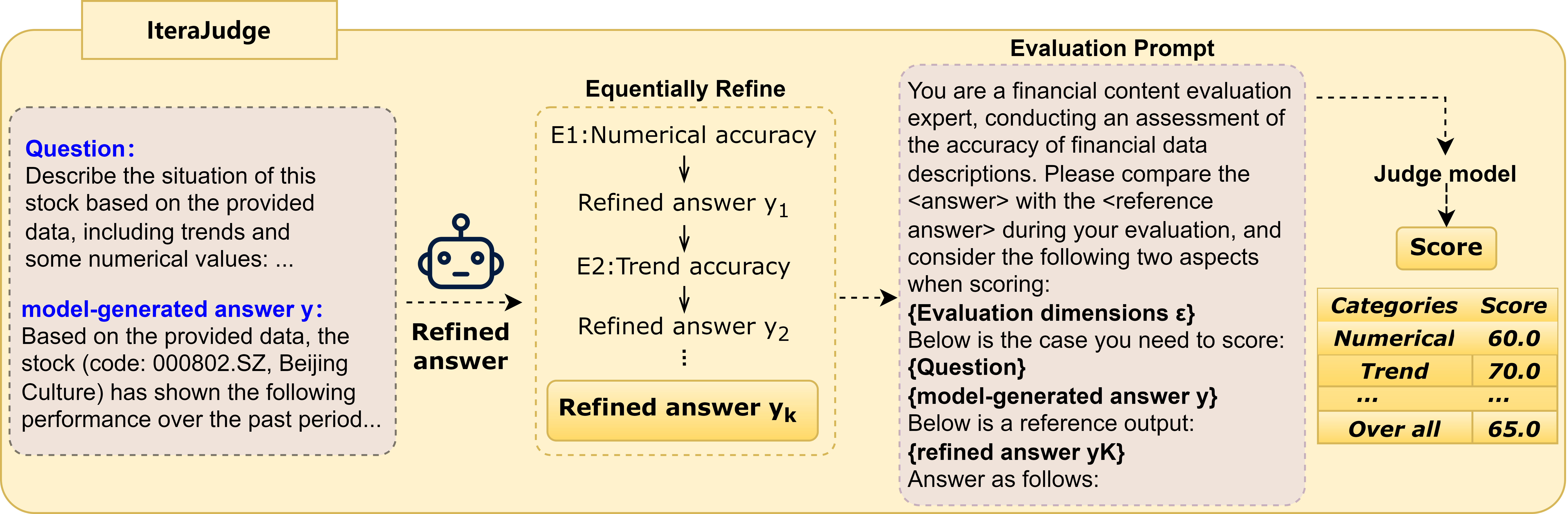}
    \caption{IteraJudge Pipeline.}
    \label{fig:oms}
\end{figure}

As shown in Figure~\ref{fig:oms}, \textbf{\TheEvaName{}} evaluation framework performs dimension-decoupled assessment through a three-phase pipeline: 

\begin{enumerate}[leftmargin=*,noitemsep]
    \item Given question \(q\) and initial answer \(y \sim p_{\text{model}}(\cdot|q)\), we sequentially refine the output across dimensions \(\mathcal{E} = \{e_1,\ldots,e_K\}\) via prompted LLM transformations:
    \begin{equation}
        y_k = \text{LLM}_{\text{refine}}(y_{k - 1} \parallel \mathcal{P}(e_k,q)), \quad \text{where } y_0 = y
    \end{equation}
    creating an interpretable improvement trajectory \(\{y_k\}_{k = 0}^K\).
    \item The fully refined \(y_K\) serves as an auto - generated quality benchmark.
    \item A judge model computes the final score through contrastive evaluation:
    \begin{equation}
        \text{score}(y) = \text{LLM}_{\text{judge}}(q, y, y_K, \mathcal{E})
    \end{equation}
    where the delta \((y_K - y)\) quantitatively reveals the dimensional deficiencies of LLMs.
\end{enumerate}
This \textit{question-anchored}, iterative refinement process enables granular diagnosis while maintaining contextual consistency through explicit $q$-preservation in all steps.

\section{Experiments}
\label{Experiments}
This section summarizes the evaluated models (Section~\ref{sec:Evaluated Models}), including SOTA LLMs, inference-optimized models, and multimodal large language models. Section~\ref{sec:Experiment Setting} describes the experimental setup. Section~\ref{sec:Main Results} presents key results across financial tasks, followed by the performance analysis of \TheEvaName{} in Section~\ref{sec:Ablation Ecperiments}.

\subsection{Evaluated Models}
\label{sec:Evaluated Models}
We conducted a systematic evaluation of current mainstream LLMs on BizFinBench. For closed-source models, we selected five industry-recognized SOTA models: OpenAI's GPT-4o, o3 and o4-mini, Google's Gemini-2.0-Flash, and Anthropic's Claude-3.5-Sonnet. For open-source models, our evaluation covered both general-purpose LLMs including the Qwen2.5 series, Llama-3.1 series and Llama-4-Scout, as well as the financial-specialized Xuanyuan3-70B model. To comprehensively assess model capability boundaries, we also incorporated the DeepSeek-R1 series (including the R1-distill variant) which excels at complex reasoning tasks, the newly open-sourced reasoning model QwQ-32B and the recently released Qwen3 series with hybrid reasoning capabilities. Furthermore, to evaluate MLLMs on our benchmark, we extended our experiments to assess the performance of the Qwen-VL series of MLLMs\footnote{The specific details of the relevant models can be found in the Appendix~\ref{sec:Model detail}.}.

\begin{table*}[h]
  \centering
  \caption{Performance Comparison of Large Language Models on \TheName{}. The models are evaluated across multiple tasks, with results color-coded to represent the top three performers for each task: \colorbox{golden}{golden} indicates the top-performing model, \colorbox{lightblue}{silver} represents the second-best result, and \colorbox{lightgreen}{bronze} denotes the third-best performance.}
  \resizebox{\textwidth}{!}{%
    \begin{tabular}{lrrrrrrrrrr}
    \toprule
    \multicolumn{1}{c}{Model} & \multicolumn{1}{c}{AEA} & \multicolumn{1}{c}{FNC} & \multicolumn{1}{c}{FTR} & \multicolumn{1}{c}{FTU} & \multicolumn{1}{c}{FQA} & \multicolumn{1}{c}{FDD} & \multicolumn{1}{c}{ER} & \multicolumn{1}{c}{SP} & \multicolumn{1}{c}{FNER}  & \multicolumn{1}{c}{Average}\\
    \midrule
    \multicolumn{10}{c}{Propretary LLMs} \\
    ChatGPT-o3 & \cellcolor{lightblue}86.23 & 61.30     & \cellcolor{lightblue}75.36      & \cellcolor{golden}89.15    &   \cellcolor{lightblue}91.25    &  \cellcolor{lightgreen}98.55     & \cellcolor{lightgreen}44.48     & 53.27     & 65.13    & \cellcolor{golden}73.86 \\
    ChatGPT-o4-mini & \cellcolor{lightgreen}85.62 & 60.10     & 71.23      & 74.40    &   90.27    &  95.73     & \cellcolor{golden}47.67     & 52.32     & 64.24    & 71.29 \\
    GPT-4o & 79.42 & 56.51     & \cellcolor{golden}76.20      &   82.37    &   87.79    &  \cellcolor{golden}98.84     & \cellcolor{lightblue}45.33     & 54.33     & 65.37    & \cellcolor{lightgreen}71.80 \\
    Gemini-2.0-Flash & \cellcolor{golden}86.94 & \cellcolor{lightgreen}62.67    & 73.97    &   82.55   &  90.29    & \cellcolor{lightblue}98.62   & 22.17     & \cellcolor{lightgreen}56.14     & 54.43    & 69.75 \\
    Claude-3.5-Sonnet & 84.68 & \cellcolor{lightblue}63.18    & 42.81     &  \cellcolor{lightblue}88.05     &  87.35     &  96.85     & 16.67     & 47.60     & 63.09    & 65.59 \\
    \midrule
    \multicolumn{10}{c}{Open source LLMs} \\
    Qwen2.5-7B-Instruct & 73.87 & 32.88     & 39.38     &  79.03     &  83.34     &   78.93    & 37.50     & 51.91     & 30.31    & 56.35 \\
    Qwen2.5-72B-Instruct & 69.27 & 54.28    & 70.72     &   85.29    &  87.79     & 97.43    & 35.33     & 55.13     & 54.02    & 67.70 \\
    Qwen2.5-VL-3B & 53.85 & 15.92     & 17.29     &  8.95     &  81.60     &  59.44    & 39.50     & 52.49     & 21.57    & 38.96 \\
    Qwen2.5-VL-7B & 73.87 & 32.71     &  40.24    &  77.85     &  83.94     &  77.41    & 38.83     & 51.91     & 33.40    & 56.68 \\
    Qwen2.5-VL-14B & 37.12   & 41.44     & 53.08     &   82.07    &  84.23     &  7.97    & 37.33     & 54.93     & 47.47    & 49.52 \\
    Qwen2.5-VL-32B & 76.79 & 50.00     & 62.16     &  83.57     &  85.30     &  95.95    & 40.50     & 54.93     & \cellcolor{lightgreen}68.36    & 68.62 \\
    Qwen2.5-VL-72B & 69.55 & 54.11     & 69.86     &  85.18     &  87.37     &  97.34    & 35.00     & 54.94     & 54.41    & 67.53 \\
    Qwen3-1.7B & 77.40 & 35.80 & 33.40 & 75.82 & 73.81 & 78.62 & 22.40 & 48.53 & 11.23 & 50.78 \\
    Qwen3-4B & 83.60 & 47.40 & 50.00 & 78.19 & 82.24 & 80.16 & 42.20 & 50.51 & 25.19 & 59.94 \\
    Qwen3-14B & 84.20 & 58.20 & 65.80 & 82.19 & 84.12 & 92.91 & 33.00 & 52.31 & 50.70 & 67.05 \\
    Qwen3-32B & 83.80 & 59.60 & 64.60 & 85.12 & 85.43 & 95.37 & 39.00 & 52.26 & 49.19 & 68.26 \\
    Xuanyuan3-70B & 12.14 & 19.69     & 15.41     &  80.89  &  86.51     &  83.90     & 29.83     & 52.62     & 37.33    & 46.48 \\
    Llama-3.1-8B-Instruct & 73.12 & 22.09    & 2.91      &  77.42     &  76.18     &  69.09     & 29.00     & 54.21     & 36.56    & 48.95 \\
    Llama-3.1-70B-Instruct & 16.26 & 34.25    & 56.34     &  80.64     &   79.97    &  86.90     & 33.33     & \cellcolor{golden}62.16     & 45.95    & 55.09 \\
    Llama 4 Scout & 73.60 & 45.80 & 44.20 & 85.02 & 85.21 & 92.32 & 25.60 & 55.76 & 43.00 & 61.17 \\
    DeepSeek-V3 (671B) & 74.34 & 61.82    & 72.60     &  \cellcolor{lightblue}86.54     &  \cellcolor{lightgreen}91.07     & 98.11      & 32.67     & 55.73     & \cellcolor{lightblue}71.24    & 71.57 \\
    DeepSeek-R1 (671B) & 80.36 & \cellcolor{golden}64.04   & \cellcolor{lightgreen}75.00     &  81.96     &  \cellcolor{golden}91.44     & 98.41      & 39.67     & 55.13     &  \cellcolor{golden}71.46   & \cellcolor{lightblue}73.05 \\
    QwQ-32B & 84.02 & 52.91    & 64.90     &  84.81     &  89.60     &  94.20     & 34.50     & \cellcolor{lightblue}56.68     & 30.27    & 65.77 \\
    DeepSeek-R1-Distill-Qwen-14B & 71.33 & 44.35 & 16.95     &   81.96    &  85.52     & 92.81   & 39.50     & 50.20     & 52.76    & 59.49 \\
    DeepSeek-R1-Distill-Qwen-32B & 73.68 & 51.20 & 50.86     &  83.27     &  87.54     &  97.81     & 41.50     & 53.92     & 56.80    & 66.29 \\
    \bottomrule
    \end{tabular}%
    }
  \label{resultstable}%
\end{table*}%

\subsection{Experiment Setting}
\label{sec:Experiment Setting}
All LLMs were configured with a maximum generation length of 1,024 tokens, temperature parameter $T=0$, and batch size $B=1000$. We employed GPT-4o as the unified evaluation judge. Open-source models were deployed on an 8$\times$NVIDIA H100 cluster, while closed-source models were accessed via their official APIs. The complete evaluation required approximately 10 hours with a total computational cost of \$21,000.

To ensure standardized outputs and facilitate automated assessment, we constrained all models to produce strictly JSON-formatted responses containing two mandatory fields:
\ding{172} Chain-of-Thought(cot): Detailed logic trace with intermediate steps;
and \ding{173} Answer: Final conclusion derived after reasoning\footnote{The formatting and style for each dataset are presented in Appendix~\ref{sec:appendixDataset example}. Appendix~\ref{Instruction} provides the evaluation prompts used in our experiments.}.

\subsection{Main Results}
\label{sec:Main Results}

Our evaluation on the \TheName{} benchmark reveals distinct capabilities of LLMs in the financial domain. All results as shown in Table~\ref{resultstable}. In the AEA task, Gemini-2.0-Flash achieves SOTA performance with a score of 86.94, closely followed by ChatGPT-o3 (86.23) and ChatGPT-o4-mini (85.62), demonstrating the strong and consistent performance of closed-source models in complex financial understanding. Moreover, proprietary models dominate knowledge-intensive tasks—exemplified by the leading performance of GPT-4o in FDD with a score of 98.84 and the leadership of ChatGPT-o3 in FTU with 89.15. However, open-source models like DeepSeek-V3 (671B) show impressive competitiveness, particularly surpassing GPT-4o (65.37) in FNER with a score of 71.46.

\begin{table*}[h]
  \centering
  \caption{Comparative Evaluation of Judgment Methods Across Different LLM Judges}
  \resizebox{0.8\textwidth}{!}{
    \begin{tabular}{lrrr}
    \toprule
    Methods & Financial Data Description & Financial Tool Usage \\
    \midrule
    LLM as a judge (GPT-4o) & Spearman: 0.4848 & Spearman: 0.8000 \\
    Ours (GPT-4o) & Spearman: 0.5684 & Spearman: 0.8667 \\
    \midrule
    LLM as a judge (DeepSeek-V3) & Spearman: 0.4685 & Spearman: 0.7500 \\
    Ours (DeepSeek-V3) & Spearman: 0.4830 & Spearman: 0.7833 \\
    \midrule
    LLM as a judge (Gemini-2.0-Flash) & Spearman: 0.3763 & Spearman: 0.7333 \\
    Ours (Gemini-2.0-Flash) & Spearman: 0.4087 & Spearman: 0.8167 \\
    \midrule
    LLM as a judge (Qwen2.5-72B-Instruct) & Spearman: 0.3112 & Spearman: 0.7000 \\
    Ours (Qwen2.5-72B-Instruct) & Spearman: 0.4282 & Spearman: 0.7500 \\
    \bottomrule
    \end{tabular}%
    }
  \label{ablation_experiments_result}%
\vspace{-4mm}
\end{table*}%

Our evaluation leads to three key insights: First, model scale plays a crucial role in numerical reasoning. Taking the Qwen3 series as an example, performance consistently improves with increasing model size, e.g., from the smallest 1.7B to the largest 32B, across nearly all tasks.
Second, FTR emerges as a particularly challenging task, with a substantial score gap of 32.19 points between the top performer, GPT-4o (76.20), and lower-performing models like Llama-3.1-8B-Instruct (2.91), highlighting the need for targeted optimization in temporal reasoning. Third, while most models excel in structured data tasks—achieving scores above 90 in cases like FDD. They underperform in more complex scenarios such as ER. Even GPT-4o scores only 45.33 in ER, with the best open-source model reaching just 41.50, underscoring significant room for improvement in financial sentiment analysis.

Additionally, we identify two intriguing phenomena: (1) Llama-3.1-70B-Instruct demonstrates strong performance in Stock Prediction (62.16) but struggles with related reasoning tasks, and (2) distilled models (e.g., DeepSeek-R1-Distill-Qwen-32B) maintain competitive performance in FTU(83.27) but exhibit limitations in temporal reasoning (FTR, 50.86), highlighting considerable variation in knowledge retention across different capabilities. These findings suggest that while current LLMs are competent at handling basic financial tasks, they still face significant limitations when dealing with complex challenges that require integrated knowledge, particularly in cross-concept reasoning within financial contexts.

\subsection{\TheEvaName{} Ablation Experiments}
\label{sec:Ablation Ecperiments}
To rigorously validate the effectiveness of \TheEvaName{}, we conducted ablation experiments on the FDD and FTU benchmark datasets. We selected Qwen2.5-7B-Instruct as the evaluated model and employed GPT-4o, DeepSeek-V3, Gemini-1.5-Flash, and Qwen2.5-72B-Instruct as judge models to evaluate its generated responses. Three sets of experiments were designed: (1) expert evaluation, (2) the vanilla LLM-as-a-Judge approach, and (3) the full \TheEvaName{} framework. We take the Spearman correlation between the evaluation methods, i.e., vanilla LLM-as-a-Judge and \TheEvaName{}.

The experimental results are demonstrated in Table~\ref{ablation_experiments_result}, compared to the LLM-as-a-Judge approach, \TheEvaName{} achieves a maximum improvement of 17.24\% and a minimum improvement of 3.09\% in terms of Spearman correlation on the FDD benchmark dataset. In the FTU benchmark dataset, it shows a maximum improvement of 11.37\% and a minimum improvement of 4.44\%. These results confirm the effectiveness of \TheEvaName{} in mitigating evaluation bias.

\section{Conclusion}
\label{conclusion}
In this work, we propose \TheName{}, i.e., the first open-source benchmark dataset, which consists of the dataset deeply integrated with real-world financial business scenarios and a iterative calibration-based evaluation framework, i.e., \TheEvaName{}. We conducted a comprehensive evaluation of 25 SOTA LLMs, encompassing both closed-source and open-source models, across multiple task dimensions. Our results reveal significant performance gaps between existing LLMs and human-level expectations in several business-critical areas, highlighting the unique challenges of financial artificial intelligence. We find that no model dominates every task, while ChatGPT-o3, ChatGPT-o4-mini, GPT-4o, Gemini-2.0-Flash, DeepSeek-R1, and Llama-3.1-70B-Instruct corresponds to the best performance in diverse metrics. In addition, experimental results also demonstrate that closed-source models place in the top three on eight of nine subtasks. Furthermore, extensive experimental results reveal significant advantages of \TheEvaName{}. \TheName{} serves not only as a rigorous benchmark for evaluating financial reasoning capabilities, but also as a practical guide for deploying LLMs in real-world financial applications. We believe this benchmark can accelerate progress in the development of trustworthy, high-performing financial language models.


\bibliographystyle{unsrtnat}
\bibliography{Ref}

\begin{thebibliography}{36}
\providecommand{\natexlab}[1]{#1}
\providecommand{\url}[1]{\texttt{#1}}
\expandafter\ifx\csname urlstyle\endcsname\relax
  \providecommand{\doi}[1]{doi: #1}\else
  \providecommand{\doi}{doi: \begingroup \urlstyle{rm}\Url}\fi

\bibitem[Chen et~al.(2024)Chen, Ma, Zhang, Hao, Yan, Nourbakhsh, Yang, McAuley, Petzold, and Wang]{chen2024survey}
Zhiyu~Zoey Chen, Jing Ma, Xinlu Zhang, Nan Hao, An~Yan, Armineh Nourbakhsh, Xianjun Yang, Julian McAuley, Linda Petzold, and William~Yang Wang.
\newblock A survey on large language models for critical societal domains: Finance, healthcare, and law.
\newblock \emph{arXiv preprint arXiv:2405.01769}, 2024.

\bibitem[Liu et~al.(2025)Liu, Zhang, Zhang, Zhang, Gong, Li, Lu, Zhou, Lu, Gan, et~al.]{hithinkliu2025nexus}
Che Liu, Yingji Zhang, Dong Zhang, Weijie Zhang, Chenggong Gong, Haohan Li, Yu~Lu, Shilin Zhou, Yue Lu, Ziliang Gan, et~al.
\newblock Nexus-o: An omni-perceptive and-interactive model for language, audio, and vision.
\newblock \emph{arXiv preprint arXiv:2503.01879}, 2025.

\bibitem[Zhang et~al.(2023{\natexlab{a}})Zhang, Wu, Chen, Wen, Nepal, Paris, and Xiang]{hithinkzhang2023dynalogue}
Rongjunchen Zhang, Tingmin Wu, Xiao Chen, Sheng Wen, Surya Nepal, Cecile Paris, and Yang Xiang.
\newblock Dynalogue: A transformer-based dialogue system with dynamic attention.
\newblock In \emph{Proceedings of the ACM Web Conference 2023}, pages 1604--1615, 2023{\natexlab{a}}.

\bibitem[Lu et~al.(2024)Lu, Ju, Chen, Pei, and Cai]{lu2024grace}
Guilong Lu, Xiaolin Ju, Xiang Chen, Wenlong Pei, and Zhilong Cai.
\newblock Grace: Empowering llm-based software vulnerability detection with graph structure and in-context learning.
\newblock \emph{Journal of Systems and Software}, 212:\penalty0 112031, 2024.

\bibitem[Xie et~al.(2025)Xie, Chen, Chen, Peng, Hu, Lin, Peng, Huang, Zhang, Keloth, et~al.]{xie2025medical}
Qianqian Xie, Qingyu Chen, Aokun Chen, Cheng Peng, Yan Hu, Fongci Lin, Xueqing Peng, Jimin Huang, Jeffrey Zhang, Vipina Keloth, et~al.
\newblock Medical foundation large language models for comprehensive text analysis and beyond.
\newblock \emph{npj Digital Medicine}, 8\penalty0 (1):\penalty0 141, 2025.

\bibitem[Zhao et~al.(2024)Zhao, Liu, Wu, Li, Yang, Shu, Xu, Dai, Zhao, Mai, et~al.]{zhao2024revolutionizing}
Huaqin Zhao, Zhengliang Liu, Zihao Wu, Yiwei Li, Tianze Yang, Peng Shu, Shaochen Xu, Haixing Dai, Lin Zhao, Gengchen Mai, et~al.
\newblock Revolutionizing finance with llms: An overview of applications and insights.
\newblock \emph{arXiv preprint arXiv:2401.11641}, 2024.

\bibitem[Gan et~al.(2024)Gan, Lu, Zhang, Li, Liu, Liu, Liu, Wu, Fu, Xu, et~al.]{gan2024mme}
Ziliang Gan, Yu~Lu, Dong Zhang, Haohan Li, Che Liu, Jian Liu, Ji~Liu, Haipang Wu, Chaoyou Fu, Zenglin Xu, et~al.
\newblock Mme-finance: A multimodal finance benchmark for expert-level understanding and reasoning.
\newblock \emph{arXiv preprint arXiv:2411.03314}, 2024.

\bibitem[Du et~al.(2024)Du, Xing, Mao, and Cambria]{du2024evaluation}
Kelvin Du, Frank Xing, Rui Mao, and Erik Cambria.
\newblock An evaluation of reasoning capabilities of large language models in financial sentiment analysis.
\newblock In \emph{2024 IEEE Conference on Artificial Intelligence (CAI)}, pages 189--194. IEEE, 2024.

\bibitem[Zhang et~al.(2023{\natexlab{b}})Zhang, Cai, Liu, Yang, Dai, Liao, Qin, Li, Liu, Liu, Zhu, Wu, Guo, and Chen]{FinEval}
Liwen Zhang, Weige Cai, Zhaowei Liu, Zhi Yang, Wei Dai, Yujie Liao, Qianru Qin, Yifei Li, Xingyu Liu, Zhiqiang Liu, Zhoufan Zhu, Anbo Wu, Xin Guo, and Yun Chen.
\newblock Fineval: A chinese financial domain knowledge evaluation benchmark for large language models, 2023{\natexlab{b}}.

\bibitem[Wang et~al.(2024)Wang, Chen, Fu, Liao, Zhang, Wu, Yu, Xu, Zhang, Luo, et~al.]{wang2024leave}
Minzheng Wang, Longze Chen, Cheng Fu, Shengyi Liao, Xinghua Zhang, Bingli Wu, Haiyang Yu, Nan Xu, Lei Zhang, Run Luo, et~al.
\newblock Leave no document behind: Benchmarking long-context llms with extended multi-doc qa.
\newblock \emph{arXiv preprint arXiv:2406.17419}, 2024.

\bibitem[Team(2023)]{team2023FinEva}
Fin-Eva Team.
\newblock Fin-eva version 1.0, 2023.

\bibitem[Gu et~al.(2024)Gu, Jiang, Shi, Tan, Zhai, Xu, Li, Shen, Ma, Liu, et~al.]{gu2024survey}
Jiawei Gu, Xuhui Jiang, Zhichao Shi, Hexiang Tan, Xuehao Zhai, Chengjin Xu, Wei Li, Yinghan Shen, Shengjie Ma, Honghao Liu, et~al.
\newblock A survey on llm-as-a-judge.
\newblock \emph{arXiv preprint arXiv:2411.15594}, 2024.

\bibitem[Zhang et~al.(2024{\natexlab{a}})Zhang, Wang, Yu, Jiang, Wu, Li, Wang, Jiang, Shang, Tang, et~al.]{zhang2024reviseval}
Qiyuan Zhang, Yufei Wang, Tiezheng Yu, Yuxin Jiang, Chuhan Wu, Liangyou Li, Yasheng Wang, Xin Jiang, Lifeng Shang, Ruiming Tang, et~al.
\newblock Reviseval: Improving llm-as-a-judge via response-adapted references.
\newblock \emph{arXiv preprint arXiv:2410.05193}, 2024{\natexlab{a}}.

\bibitem[Shah et~al.(2022)Shah, Chawla, Eidnani, Shah, Du, Chava, Raman, Smiley, Chen, and Yang]{shah2022flue}
Raj Shah, Kunal Chawla, Dheeraj Eidnani, Agam Shah, Wendi Du, Sudheer Chava, Natraj Raman, Charese Smiley, Jiaao Chen, and Diyi Yang.
\newblock When flue meets flang: Benchmarks and large pretrained language model for financial domain.
\newblock In \emph{Proceedings of the 2022 Conference on Empirical Methods in Natural Language Processing}, pages 2322--2335, 2022.

\bibitem[Xie et~al.(2023)Xie, Han, Zhang, Lai, Peng, Lopez-Lira, and Huang]{xie2023pixiu}
Qianqian Xie, Weiguang Han, Xiao Zhang, Yanzhao Lai, Min Peng, Alejandro Lopez-Lira, and Jimin Huang.
\newblock Pixiu: A large language model, instruction data and evaluation benchmark for finance.
\newblock \emph{arXiv preprint arXiv:2306.05443}, 2023.

\bibitem[Chen et~al.(2022{\natexlab{a}})Chen, Chen, Smiley, and Sameena~Shah]{chen2022finqa}
Zhiyu Chen, Wenhu Chen, Charese Smiley, and et~al. Sameena~Shah.
\newblock Finqa: A dataset of numerical reasoning over financial data, 2022{\natexlab{a}}.

\bibitem[Chen et~al.(2022{\natexlab{b}})Chen, Li, Smiley, Ma, Shah, and Wang]{chen2022convfinqa}
Zhiyu Chen, Shiyang Li, Charese Smiley, Zhiqiang Ma, Sameena Shah, and William~Yang Wang.
\newblock Convfinqa: Exploring the chain of numerical reasoning in conversational finance question answering, 2022{\natexlab{b}}.

\bibitem[{D}uxiaoman DI~{T}eam(2023)]{financeIQ2023}
{D}uxiaoman DI~{T}eam.
\newblock {F}inance{IQ}, 2023.
\newblock URL \url{https://github.com/Duxiaoman-DI/XuanYuan/tree/main/FinanceIQ}.
\newblock Accessed: 2024-03-18.

\bibitem[Lei et~al.(2023)Lei, Li, Jiang, Hu, Cheng, Ding, and Jiang]{lei2023cfbenchmark}
Yang Lei, Jiangtong Li, Ming Jiang, Junjie Hu, Dawei Cheng, Zhijun Ding, and Changjun Jiang.
\newblock Cfbenchmark: Chinese financial assistant benchmark for large language model, 2023.

\bibitem[Chen et~al.(2023)Chen, Wang, Long, Zhang, Lu, Li, Wang, Xu, Bai, Huang, et~al.]{chen2023disc}
Wei Chen, Qiushi Wang, Zefei Long, Xianyin Zhang, Zhongtian Lu, Bingxuan Li, Siyuan Wang, Jiarong Xu, Xiang Bai, Xuanjing Huang, et~al.
\newblock Disc-finllm: A chinese financial large language model based on multiple experts fine-tuning.
\newblock \emph{arXiv preprint arXiv:2310.15205}, 2023.

\bibitem[Zhang et~al.(2023{\natexlab{c}})Zhang, Li, and Yang]{zhang2023cgce}
Xuanyu Zhang, Bingbing Li, and Qing Yang.
\newblock Cgce: A chinese generative chat evaluation benchmark for general and financial domains, 2023{\natexlab{c}}.

\bibitem[Araci(2019)]{araci2019finbert}
Dogu Araci.
\newblock Finbert: Financial sentiment analysis with pre-trained language models, 2019.

\bibitem[Yang et~al.(2023{\natexlab{a}})Yang, Tang, and Tam]{yang2023investlm}
Yi~Yang, Yixuan Tang, and Kar~Yan Tam.
\newblock Investlm: A large language model for investment using financial domain instruction tuning, 2023{\natexlab{a}}.

\bibitem[Yang et~al.(2023{\natexlab{b}})Yang, Liu, and Wang]{yang2023fingpt}
Hongyang Yang, Xiao-Yang Liu, and Christina~Dan Wang.
\newblock Fingpt: Open-source financial large language models, 2023{\natexlab{b}}.

\bibitem[Touvron et~al.(2023)Touvron, Lavril, Izacard, Martinet, Lachaux, Lacroix, Rozi{\`e}re, Goyal, Hambro, Azhar, et~al.]{touvron2023llama1}
Hugo Touvron, Thibaut Lavril, Gautier Izacard, Xavier Martinet, Marie-Anne Lachaux, Timoth{\'e}e Lacroix, Baptiste Rozi{\`e}re, Naman Goyal, Eric Hambro, Faisal Azhar, et~al.
\newblock Llama: Open and efficient foundation language models.
\newblock \emph{arXiv preprint arXiv:2302.13971}, 2023.

\bibitem[Wu et~al.(2023)Wu, Irsoy, Lu, Dabravolski, Dredze, Gehrmann, Kambadur, Rosenberg, and Mann]{wu2023bloomberggpt}
Shijie Wu, Ozan Irsoy, Steven Lu, Vadim Dabravolski, Mark Dredze, Sebastian Gehrmann, Prabhanjan Kambadur, David Rosenberg, and Gideon Mann.
\newblock Bloomberggpt: A large language model for finance, 2023.

\bibitem[Zhang et~al.(2024{\natexlab{b}})Zhang, Qiu, Feng, Li, Ma, Zhang, Ju, Yan, and Xie]{zhang2024baichuan4}
Hanyu Zhang, Boyu Qiu, Yuhao Feng, Shuqi Li, Qian Ma, Xiyuan Zhang, Qiang Ju, Dong Yan, and Jian Xie.
\newblock Baichuan4-finance technical report.
\newblock \emph{arXiv preprint arXiv:2412.15270}, 2024{\natexlab{b}}.

\bibitem[Zhu et~al.(2025)Zhu, Chen, Dou, Li, Guo, Chen, and Zhang]{zhu2025dianjin}
Jie Zhu, Qian Chen, Huaixia Dou, Junhui Li, Lifan Guo, Feng Chen, and Chi Zhang.
\newblock Dianjin-r1: Evaluating and enhancing financial reasoning in large language models.
\newblock \emph{arXiv preprint arXiv:2504.15716}, 2025.

\bibitem[OpenAI(2023)]{openai2023gpt4}
OpenAI.
\newblock Gpt-4 technical report, 2023.

\bibitem[Team et~al.(2023)Team, Anil, Borgeaud, Wu, Alayrac, Yu, Soricut, Schalkwyk, Dai, Hauth, et~al.]{team2023gemini}
Gemini Team, Rohan Anil, Sebastian Borgeaud, Yonghui Wu, Jean-Baptiste Alayrac, Jiahui Yu, Radu Soricut, Johan Schalkwyk, Andrew~M Dai, Anja Hauth, et~al.
\newblock Gemini: a family of highly capable multimodal models.
\newblock \emph{arXiv preprint arXiv:2312.11805}, 2023.

\bibitem[Claude(2024)]{claude2024sonnet}
Claude.
\newblock Claude 3.5 sonnet, 2024.
\newblock URL \url{https://www.anthropic.com/news/claude-3-5-sonnet}.

\bibitem[Bai et~al.(2023)Bai, Bai, and Yunfei~Chu]{qwen}
Jinze Bai, Shuai Bai, and et~al. Yunfei~Chu.
\newblock Qwen technical report.
\newblock \emph{arXiv preprint arXiv:2309.16609}, 2023.

\bibitem[Team(2024)]{xuanyuan3}
XuanYuan Team.
\newblock Xuanyuan3-70b report, September 2024.
\newblock URL \url{https://github.com/Duxiaoman-DI/XuanYuan}.

\bibitem[Dubey et~al.(2024)Dubey, Jauhri, Pandey, Kadian, Al-Dahle, Letman, Mathur, Schelten, Yang, Fan, et~al.]{dubey2024llama}
Abhimanyu Dubey, Abhinav Jauhri, Abhinav Pandey, Abhishek Kadian, Ahmad Al-Dahle, Aiesha Letman, Akhil Mathur, Alan Schelten, Amy Yang, Angela Fan, et~al.
\newblock The llama 3 herd of models.
\newblock \emph{arXiv preprint arXiv:2407.21783}, 2024.

\bibitem[Liu et~al.(2024)Liu, Feng, Xue, Wang, Wu, Lu, Zhao, Deng, Zhang, Ruan, et~al.]{liu2024deepseek}
Aixin Liu, Bei Feng, Bing Xue, Bingxuan Wang, Bochao Wu, Chengda Lu, Chenggang Zhao, Chengqi Deng, Chenyu Zhang, Chong Ruan, et~al.
\newblock Deepseek-v3 technical report.
\newblock \emph{arXiv preprint arXiv:2412.19437}, 2024.

\bibitem[Team(2025)]{qwq32b}
Qwen Team.
\newblock Qwq-32b: Embracing the power of reinforcement learning, March 2025.
\newblock URL \url{https://qwenlm.github.io/blog/qwq-32b/}.

\end{thebibliography}

\appendix
\section{Reproducibility Statement}
\label{sec: Reproducibility Statement}
To ensure the reproducibility of our results, we have made substantial efforts to provide all necessary details and materials. Specifically, Section~\ref{Data Construction} presents the complete process of dataset construction, including data collection strategies. Furthermore, the benchmark setup and evaluation procedures are thoroughly described in Section~\ref{Experiments}. All evaluation metrics are clearly defined to facilitate independent verification and replication of our experiments by the research community.

\section{Limitations}
\label{limitations}
In this work, we propose a novel benchmark and conduct a comprehensive analysis of different LLMs' capabilities in solving financial business problems. However, several limitations remain:

(1) Our method for extracting final answers from model outputs is not yet perfect. In some cases, this method fails to locate an answer, leading to reported accuracy being an approximate lower bound. Additionally, due to potential formatting differences between the extracted answers and the ground truth, we employ a rule-based approach to measure exact matches between the two, which may introduce an estimated 2\% error in our experiments.

(2) Our benchmark is primarily based on currently available financial data and task settings. Although it covers multiple key sub-tasks, some business scenarios may still be underrepresented. For example, highly specialized financial tasks such as complex derivatives pricing, risk management modeling, or decision support based on real-time market data are not yet fully reflected in our benchmark. This implies that our evaluation results may not completely capture LLMs' real-world performance in more complex financial scenarios.

(3) While we evaluate multiple SOTA LLMs under the same computational environment to ensure fairness, model performance may still be influenced by training data, inference strategies, and hyperparameter settings. Additionally, discrepancies between inference mechanisms in API-based and locally deployed models could introduce experimental biases.

(4) Our evaluation primarily focuses on models’ abilities in single-turn question answering and task completion. However, in real-world applications, financial decision-making is often a complex, multi-step process involving long-term reasoning, external tool utilization, and multi-turn interactions. The current evaluation framework does not fully cover these aspects, highlighting the need for further expansion to better reflect LLMs’ potential applications in financial business scenarios.

For future work, we plan to optimize the answer extraction method to enhance evaluation accuracy and explore more advanced metrics to mitigate errors caused by format mismatches. Additionally, we aim to expand the benchmark’s coverage by incorporating more challenging financial tasks and refining experimental settings to improve reproducibility and fairness.

\section{Model detail}
\label{sec:Model detail}
To better ensure the comprehensiveness and robustness of our evaluation, we selected a wide range of models that differ in architecture, parameter size, training objectives, and domain specialization. Table~\ref{modeldetail} presents detailed information on the 25 evaluation models used in this study. 

\begin{table*}[htbp]
  \centering
  \caption{Summary of Large Language Models Evaluated on \TheName{}. * indicates that the model is a Mixture-of-Experts (MoE) model.}
    \resizebox{\textwidth}{!}{%
    \begin{tabular}{lrrrrr}
    \toprule
    Model & \multicolumn{1}{r}{Size} & \multicolumn{1}{l}{Open source} & Evalation   & \multicolumn{1}{l}{Release date} & \multicolumn{1}{l}{Domain} \\
    \midrule
    GPT-4o~\cite{openai2023gpt4} &   --    &   \ding{55}    & API   & 01/29/2025 & General \\
    ChatGPT-o3 &   --    &   \ding{55}    & API   & 04/16/2025  & General \\
    ChatGPT-o4-mini &   --    &   \ding{55}    & API   & 04/16/2025 & General \\
    Gemini-2.0-Flash~\cite{team2023gemini} &  --     &   \ding{55}    & API   & 12/11/2024 & General \\ 
    Claude-3.5-Sonnet~\cite{claude2024sonnet} &   --    &   \ding{55}    & API   & 06/20/2024 & General \\
    Qwen2.5-Instruct~\cite{qwen} &    7B,72B   &  \ding{51}     & Local & 09/19/2024 & General \\
    Qwen2.5-VL &    3B,7B,14B,32B,72B   &  \ding{51}     & Local & 01/28/2025 & General \\
    Qwen3 &    8B,14B,32B,30B*   &  \ding{51}     & Local & 04/29/2025 & General \\
    XuanYuan3-70B-Chat~\cite{xuanyuan3} &    70B   &  \ding{51}     & Local & 09/06/2024 & Finance \\
    Llama-3.1-Instruct~\cite{dubey2024llama} &   8B,70B    &  \ding{51}     & Local & 07/24/2024 & General \\
    Llama 4&  109B*     &  \ding{51}     & Local & 04/05/2025 & General \\
    DeepSeek-V3~\cite{liu2024deepseek} &  671B*     &   \ding{51}    & Local  & 12/26/2024 & General \\ 
    DeepSeek-R1~\cite{liu2024deepseek} &  671B*     &   \ding{51}    & Local & 12/26/2024 & General \\
    QwQ-32B~\cite{qwq32b} &  32B     &   \ding{51}    & Local & 03/06/2025 & General \\
    DeepSeek-R1-Distill-Qwen-14B~\cite{liu2024deepseek} &  14B,32B     &   \ding{51}    & Local & 12/26/2024 & General \\
    \bottomrule
    \end{tabular}%
    }
  \label{modeldetail}%
\end{table*}%

\section{Instruction}
\label{Instruction}
\begin{figure*}
    \centering
    \includegraphics[scale=0.05]{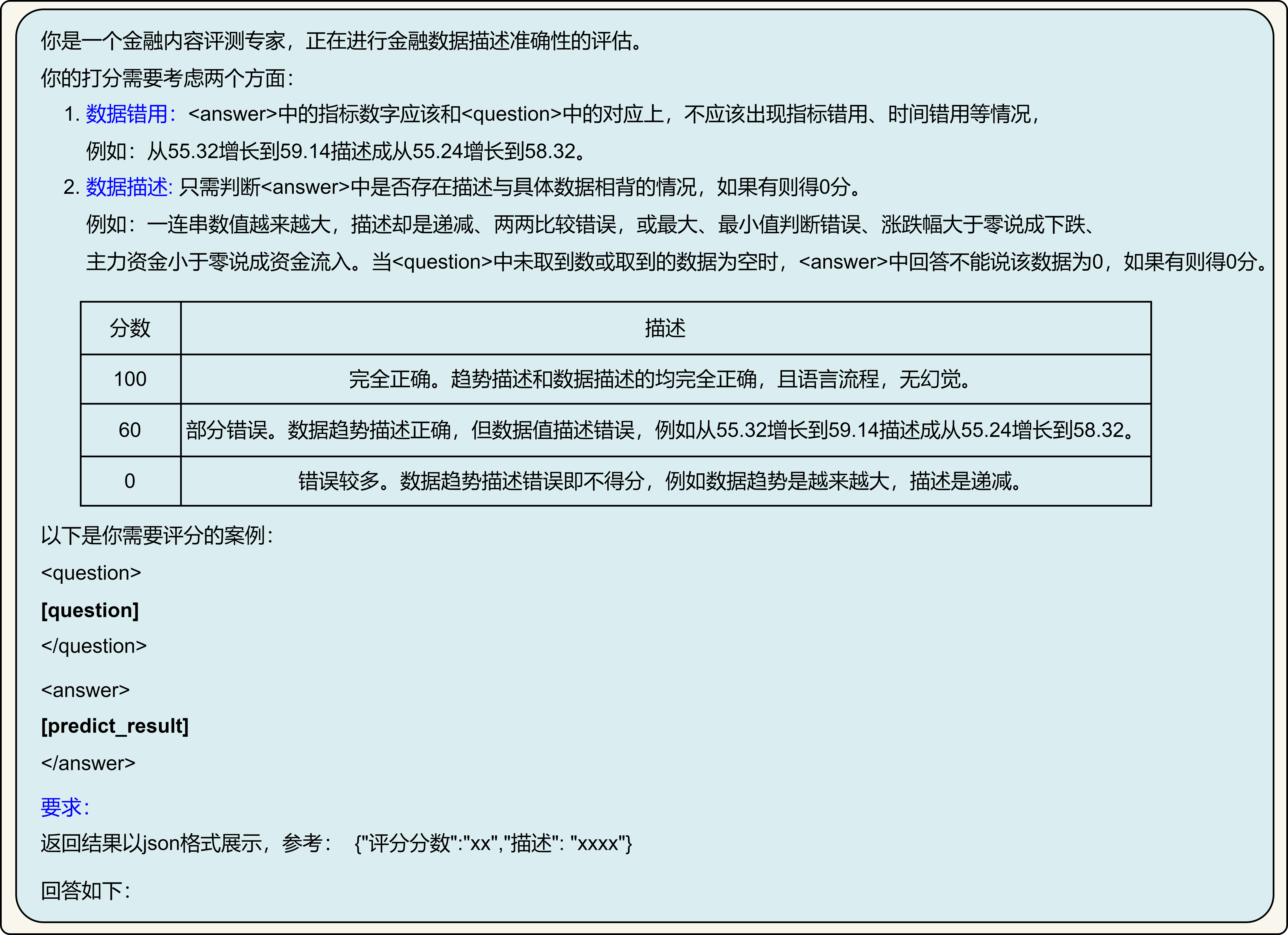}
    \caption{The instructions utilized in the evaluation of the FDD dataset.}
    \label{fig:judgepromptFDD}
\end{figure*}

\begin{figure*}
    \centering
    \includegraphics[scale=0.05]{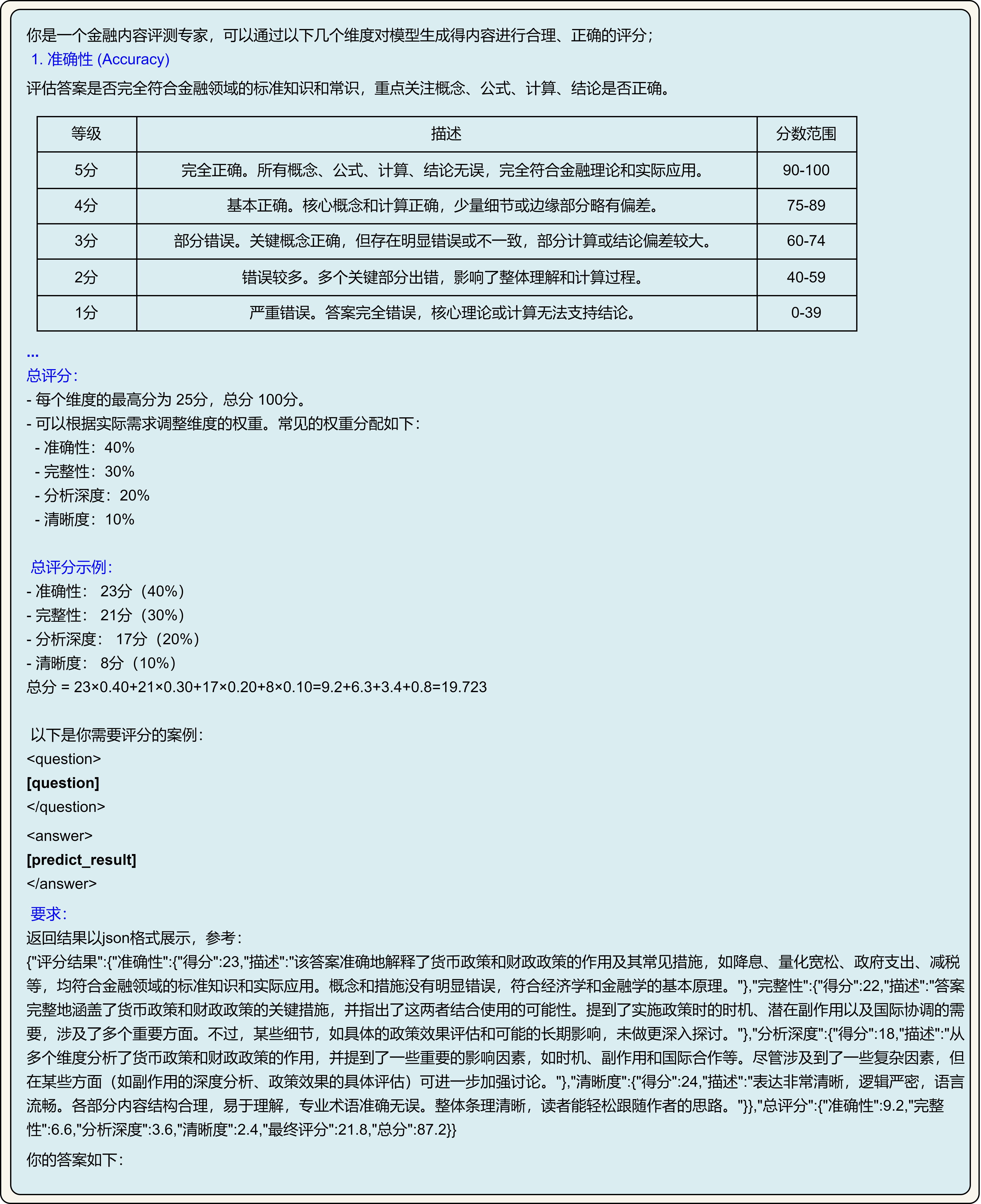}
    \caption{The instructions utilized in the evaluation of the FQA dataset.}
    \label{fig:promptFQA}
\end{figure*}

\begin{figure*}
    \centering
    \includegraphics[scale=0.05]{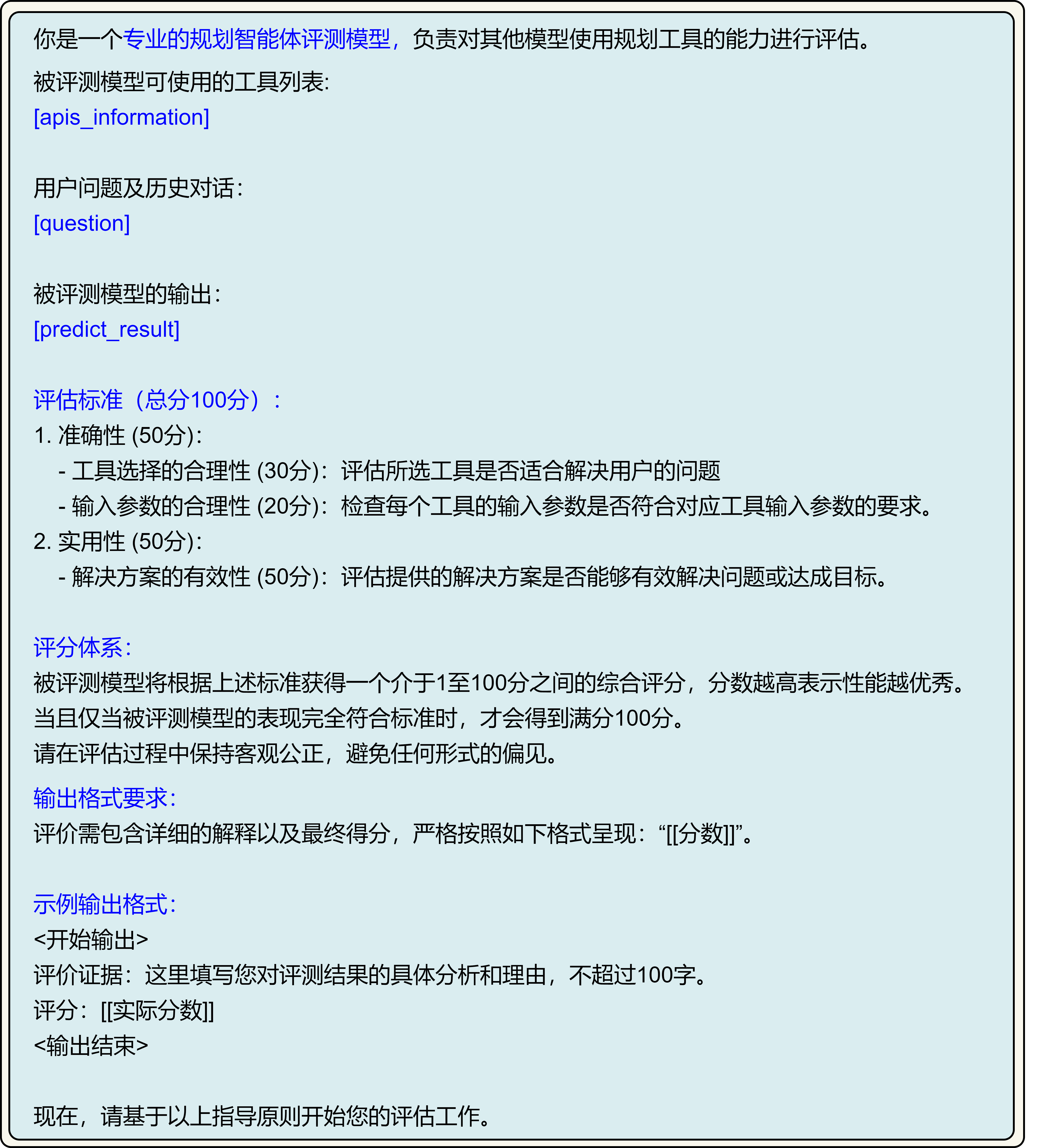}
    \caption{The instructions utilized in the evaluation of the FTU dataset.}
    \label{fig:prompttool}
\end{figure*}

Figure~\ref{fig:judgepromptFDD}, Figure~\ref{fig:promptFQA}, and Figure~\ref{fig:prompttool} illustrate the instructions used for model evaluation on the open-ended answer dataset.

\section{Dataset details}
\label{appDataset details}
The details of each dataset type are as follows. 

\begin{itemize}
    \item Anomalous Event Attribution (AEA): This dataset evaluates the model's ability to trace financial anomalies based on given information such as timestamps, news articles, financial reports, and stock movements. The model must identify the cause-and-effect relationships behind sudden market fluctuations and distinguish relevant factors from noise.
    \item Financial Numerical Computation (FNC): This dataset assesses the model's ability to perform accurate numerical calculations in financial scenarios, including interest rate calculations, return on investment (ROI), and financial ratios.
    \item Financial Time Reasoning (FTR): This dataset tests the model’s ability to understand and reason about time-based financial events, such as predicting interest accruals, identifying the impact of quarterly reports, and assessing financial trends over different periods.
    \item Financial Tool Usage (FTU): This dataset evaluates the model's ability to comprehend user queries and effectively use financial tools to solve real-world problems. It covers scenarios like investment analysis, market research, and information retrieval, requiring the model to select appropriate tools, input parameters accurately, and coordinate multiple tools when needed.
    \item Financial Knowledge QA (FQA): This dataset evaluates the model's understanding and response capabilities regarding core knowledge in the financial domain. It spans a wide range of financial topics, encompassing key areas such as fundamental financial concepts, financial markets, investment theory, macroeconomics, and finance.
    \item Financial Data Description (FDD): This dataset measures the model's ability to analyze and describe structured and unstructured financial data, such as balance sheets, stock reports, and financial statements.
    \item Emotion Recognition (ER): This dataset evaluates the model's capability to recognize nuanced user emotions in complex financial market environments. The input data encompasses multiple dimensions, including market conditions, news articles, research reports, user portfolio information, and queries. The dataset covers six distinct emotional categories: optimism, anxiety, negativity, excitement, calmness, and regret.
    \item Stock Price Prediction (SP): This dataset evaluates the model’s ability to predict future stock prices based on historical trends, financial indicators, and market news.
    \item Financial Named Entity Recognition (FNER): This dataset focuses on evaluating the model’s ability to identify and classify financial entities such as company names, stock symbols, financial instruments, regulatory agencies, and economic indicators.
\end{itemize}

Table~\ref{query_token_stats} presents their maximum token length, minimum token length, and average length.

\begin{table}[htbp]
    \centering
    \caption{Financial Datasets Query Token Length Statistics. This table presents token length statistics for queries in financial datasets, including minimum (Min), maximum (Max), average (Avg) token counts, and total query count (Count).}
    \label{query_token_stats}
    \begin{tabular}{lrrrr}
        \toprule
        \textbf{Dataset} & \textbf{Min} & \textbf{Max} & \textbf{Avg} & \textbf{Count} \\
        \midrule
        NER & 415 & 1,194 & 533.1 & 433 \\
        FTU & 4,169 & 6,289 & 4,555.5 & 641 \\
        AEA & 680 & 1,396 & 938.7 & 1,064 \\
        ER & 1,919 & 2,569 & 2,178.5 & 600 \\
        FNC & 287 & 2,698 & 650.5 & 581 \\
        FDD & 26 & 645 & 310.9 & 1,461 \\
        FTR & 203 & 8,265 & 1,162.0 & 514 \\
        FQA & 5 & 45 & 21.7 & 990 \\
        SP & 1,254 & 5,532 & 4,498.1 & 497 \\
        \bottomrule
    \end{tabular}
\end{table}

\section{Dataset example}
\label{sec:appendixDataset example}


\begin{figure*}
    \centering
    \includegraphics[width=1\linewidth]{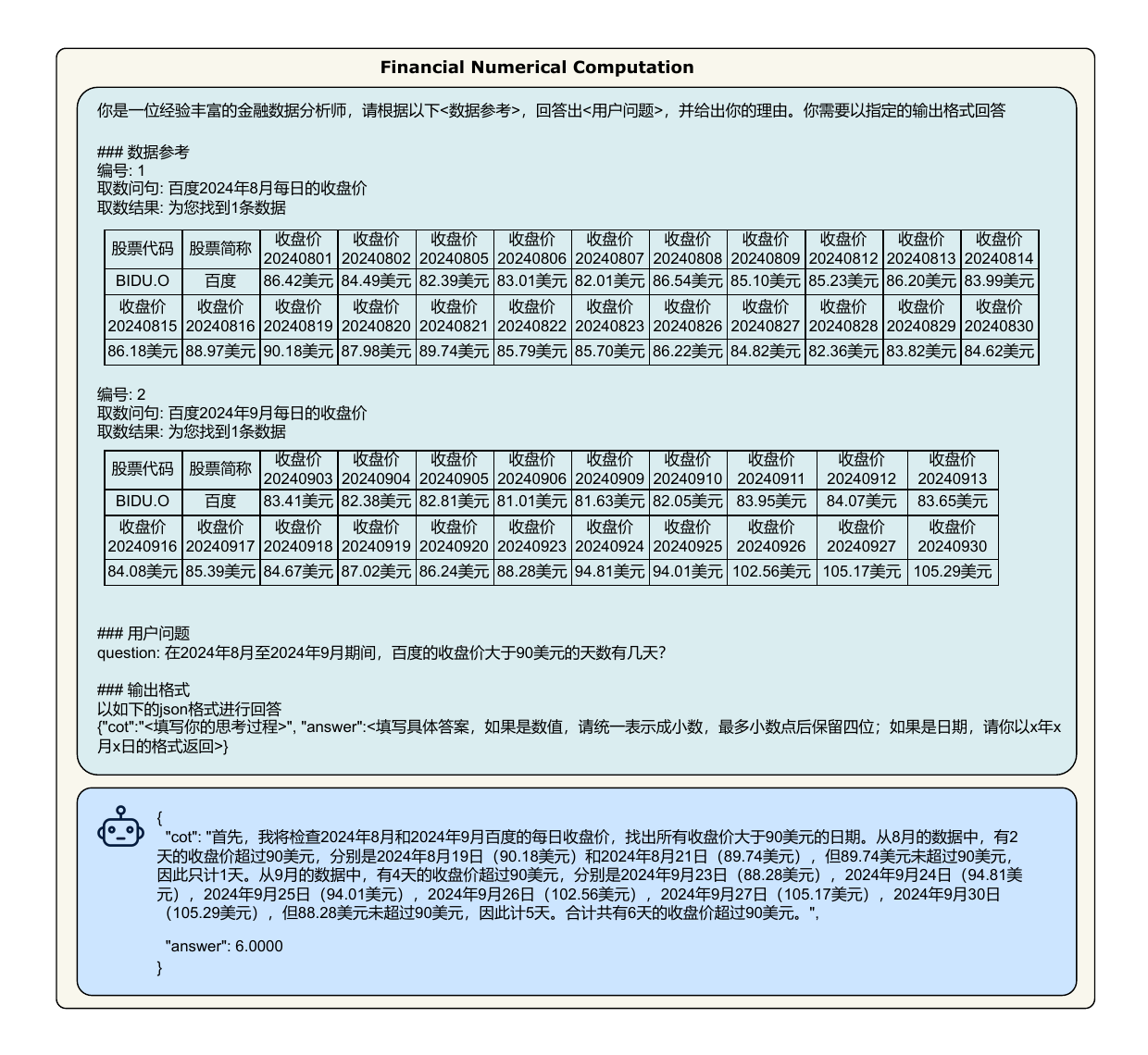}
    \caption{An example instance from the Financial Numerical Computation dataset, in which the data table is formatted using Markdown syntax.}
    \label{fig:Financial Numerical Computation}
\end{figure*}

\begin{figure*} 
    \centering
    \includegraphics[scale=0.05]{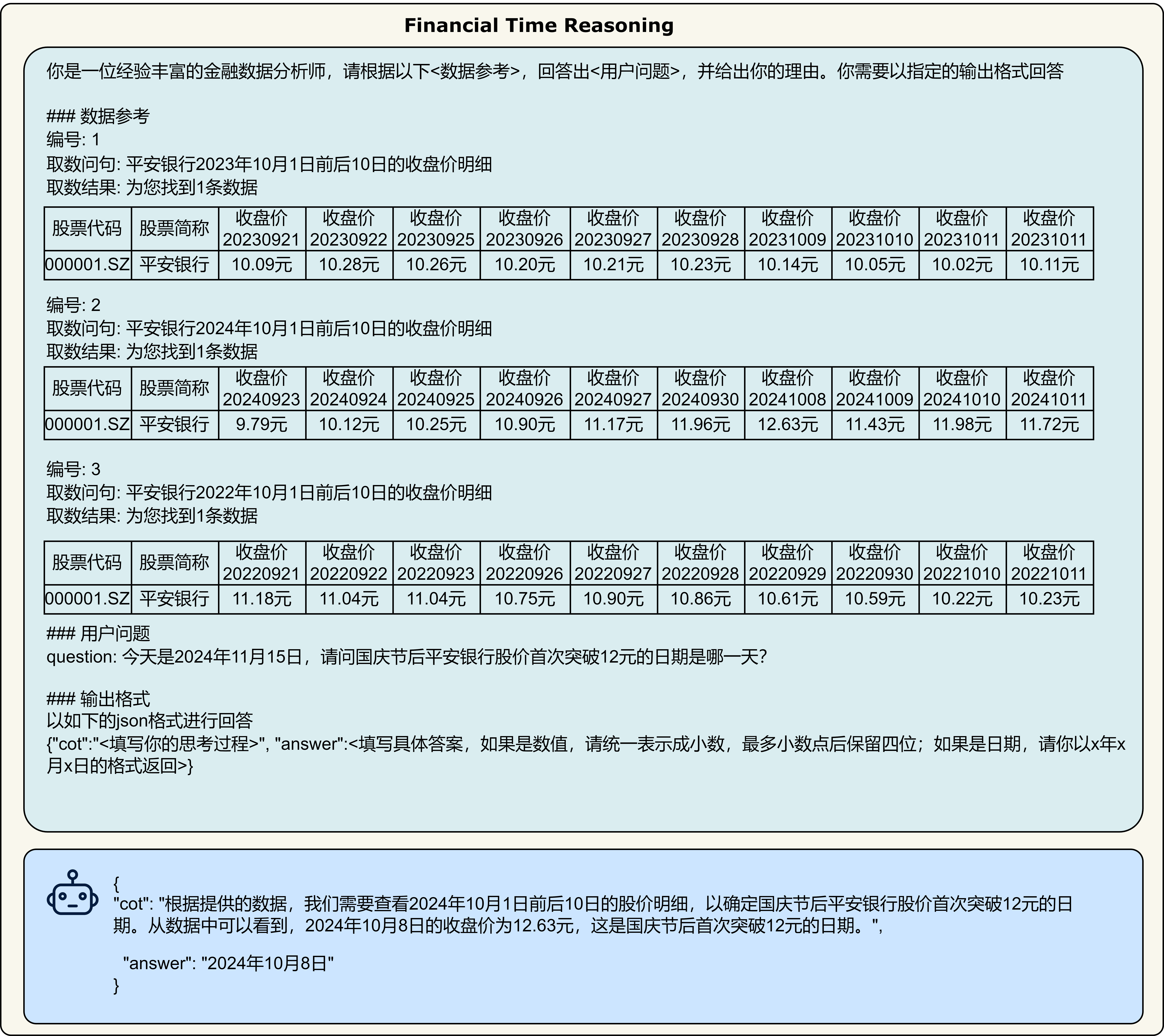}
    \caption{An example instance from the Financial Time Reasoning.}
    \label{fig:Financial Time Reasoning}
\end{figure*}

\begin{figure*}
    \centering
    \includegraphics[width=1\linewidth]{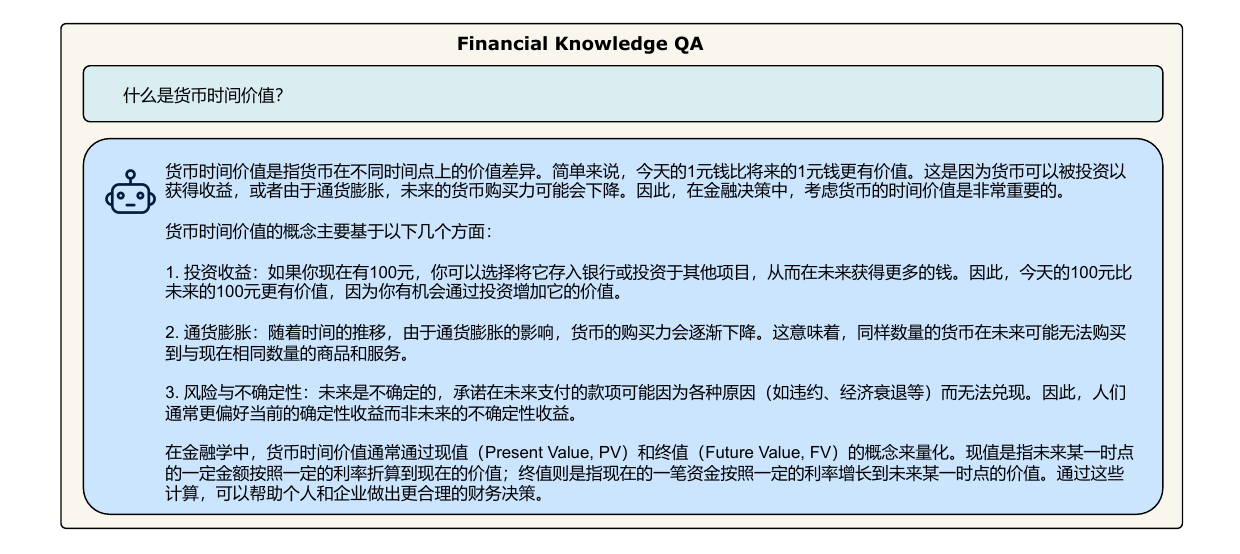}
    \caption{An example instance from the Financial Knowledge QA dataset.}
    \label{fig:Financial Knowledge QA}
\end{figure*}


\begin{figure*}
    \centering
    \includegraphics[width=1\linewidth]{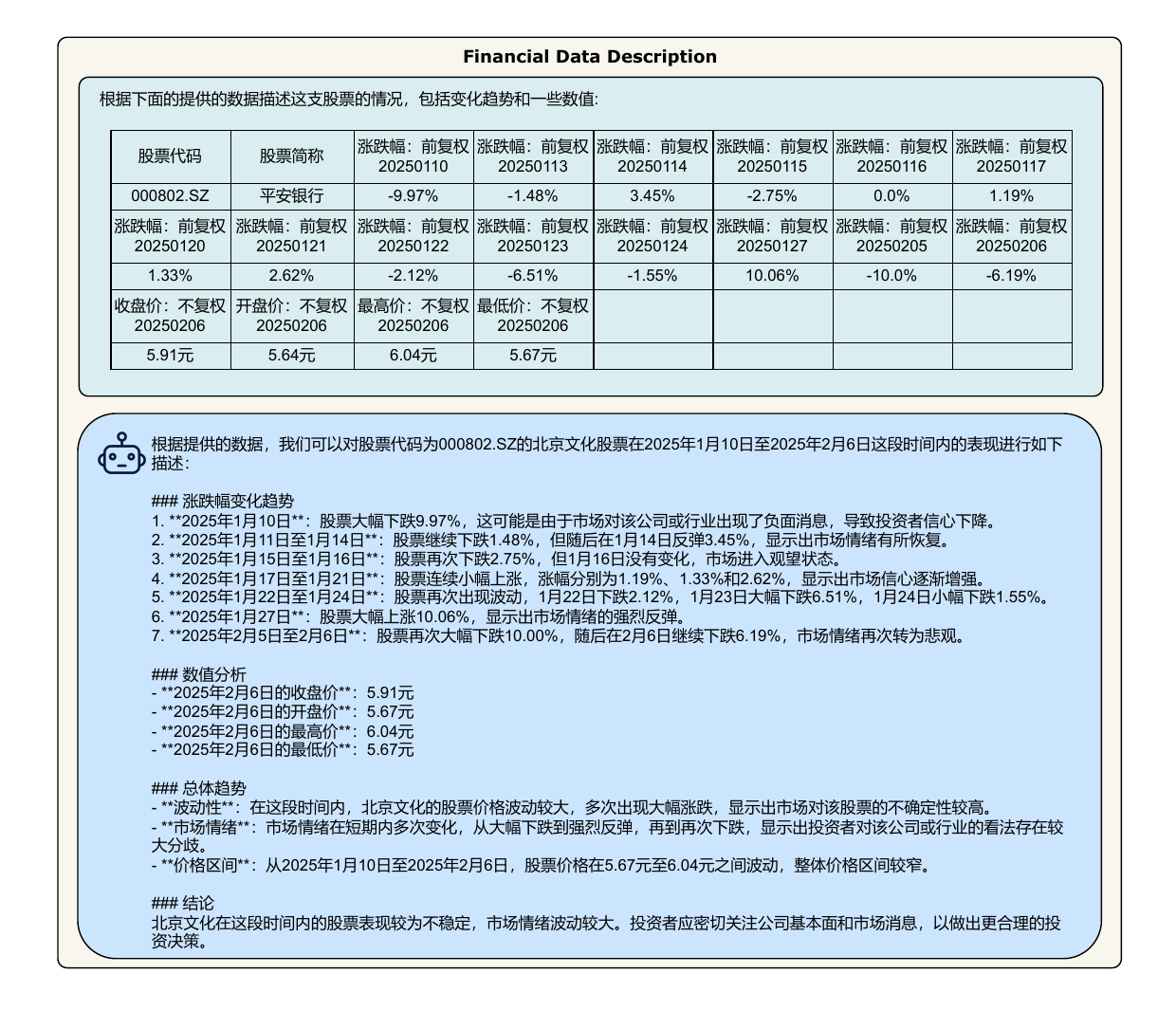}
    \caption{An example instance from the Financial Data Description.}
    \label{fig:Financial Data Description}
\end{figure*}


\begin{figure*}
    \centering
    \includegraphics[width=1\linewidth]{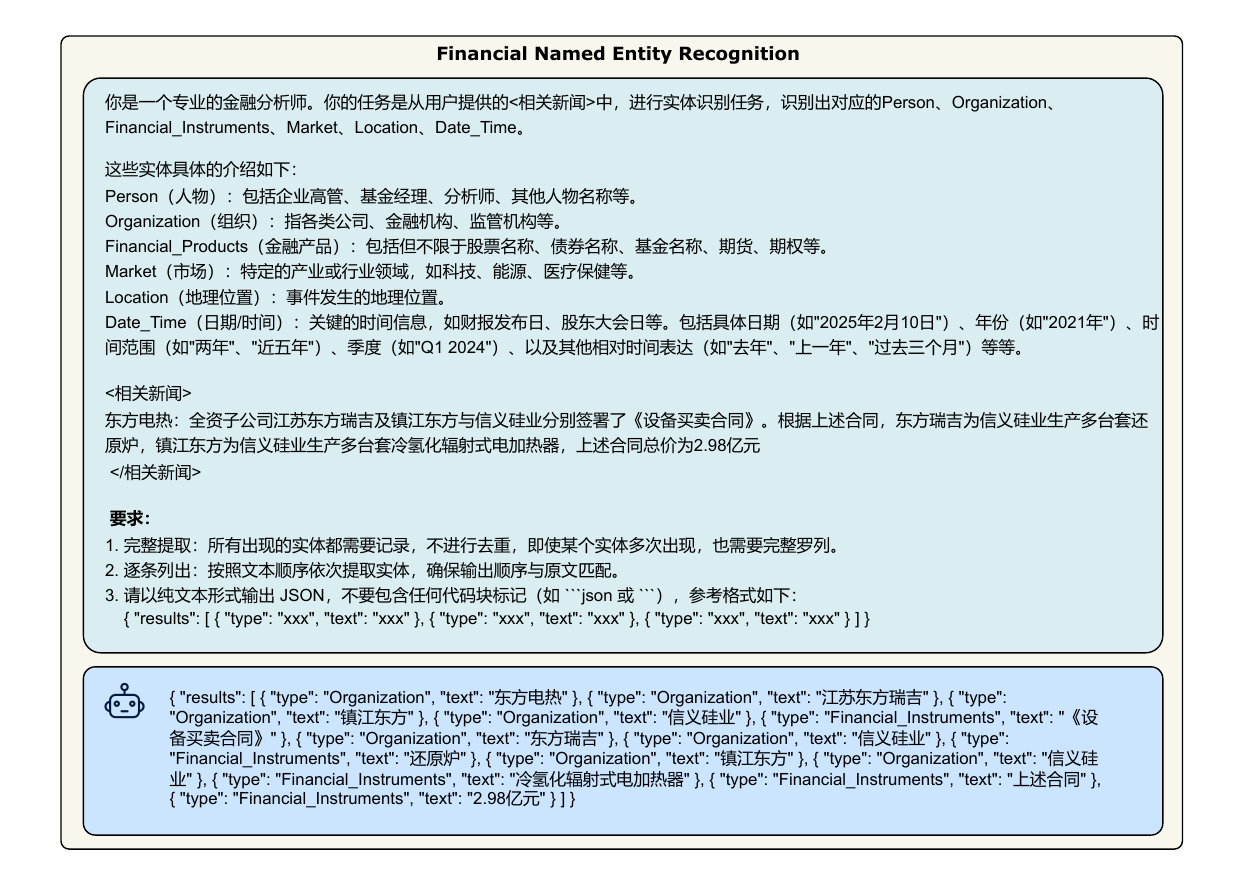}
    \caption{An example instance from the Financial Named Entity Recognition.}
    \label{fig:Financial Named Entity Recognition}
\end{figure*}

\begin{figure*}
    \centering
    \includegraphics[scale=0.05]{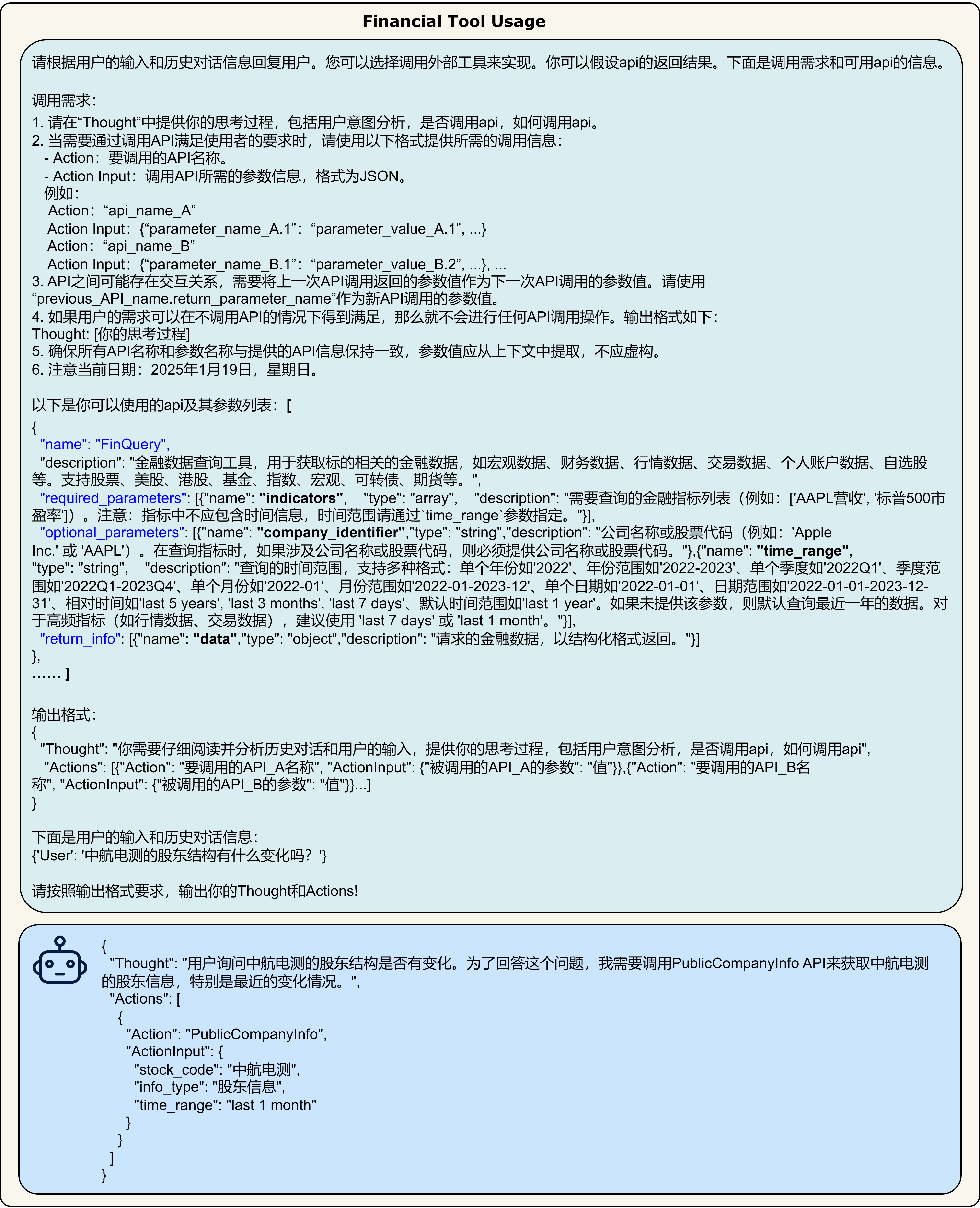}
    \caption{An example instance from the Financial Tool Usage.}
    \label{fig:Financial Tool Usage}
\end{figure*}


\begin{figure*}
    \centering
    \includegraphics[width=1\linewidth]{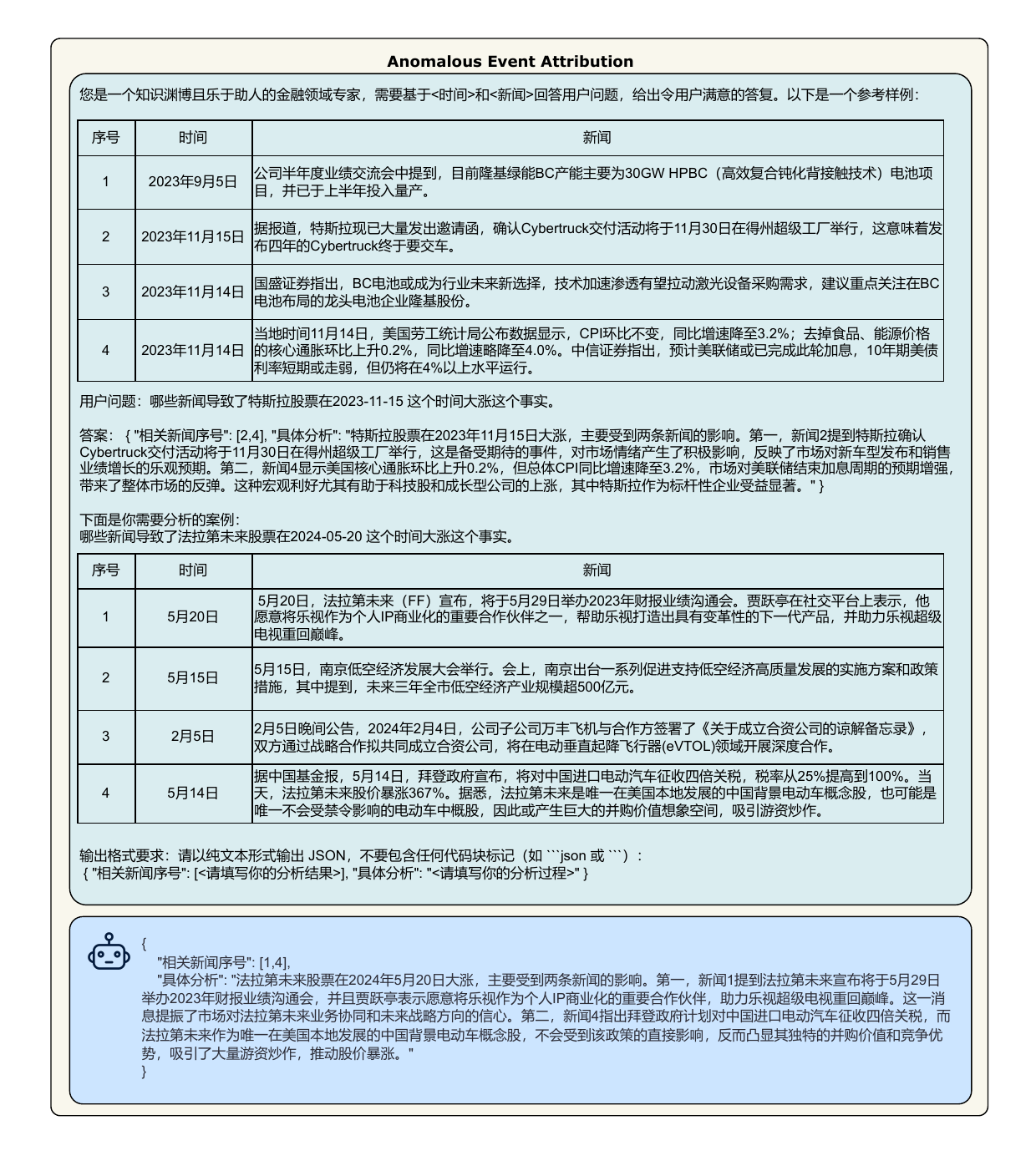}
    \caption{An example instance from the Anomalous Event Attribution.}
    \label{fig:Anomalous Event Attribution}
\end{figure*}

\begin{figure*}
    \centering
    \includegraphics[width=1\linewidth]{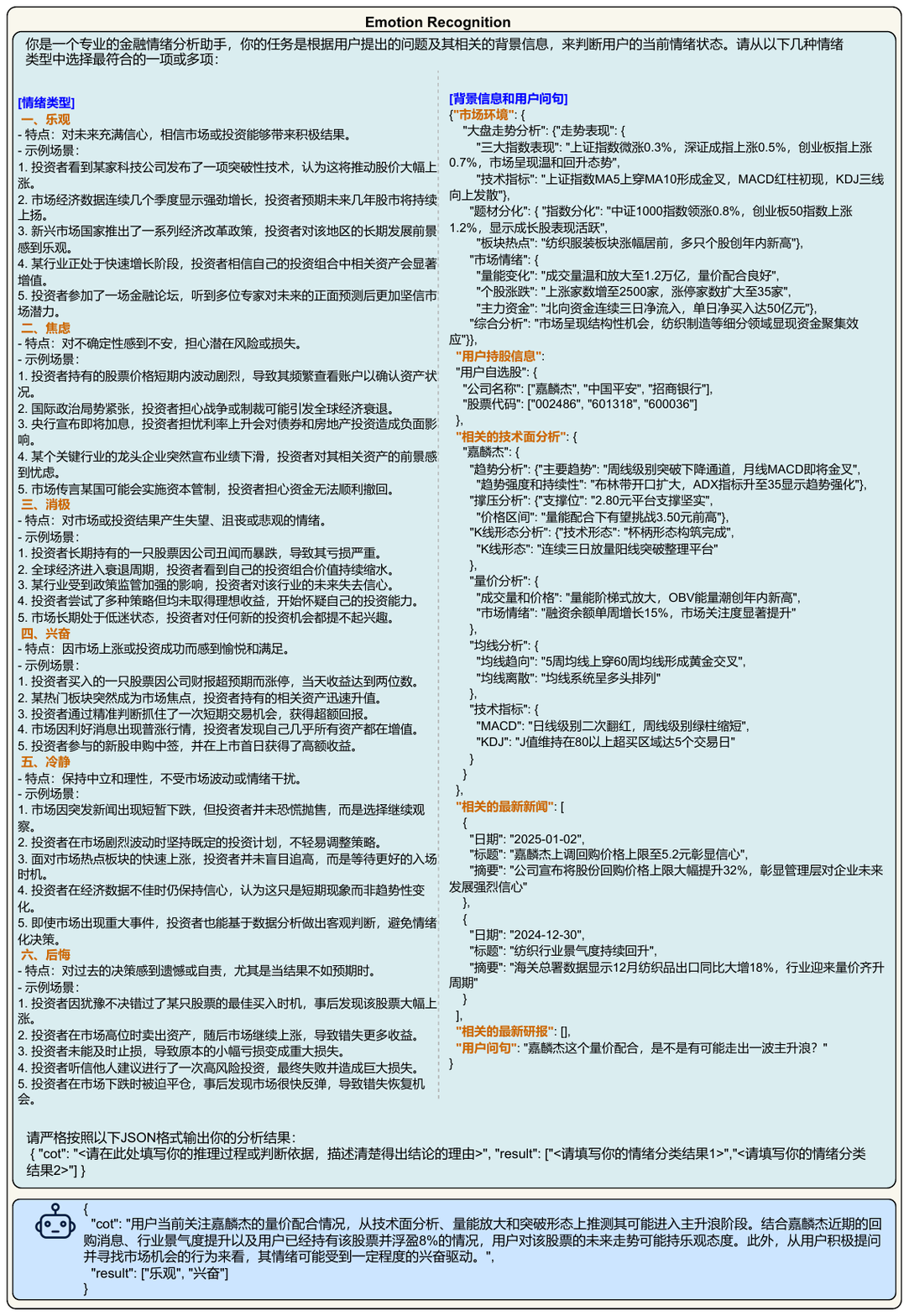}
    \caption{An example instance from the Emotion Recognition.}
    \label{fig:Emotion Recognition}
\end{figure*}

\begin{figure*}
    \centering
    \includegraphics[scale=0.05]{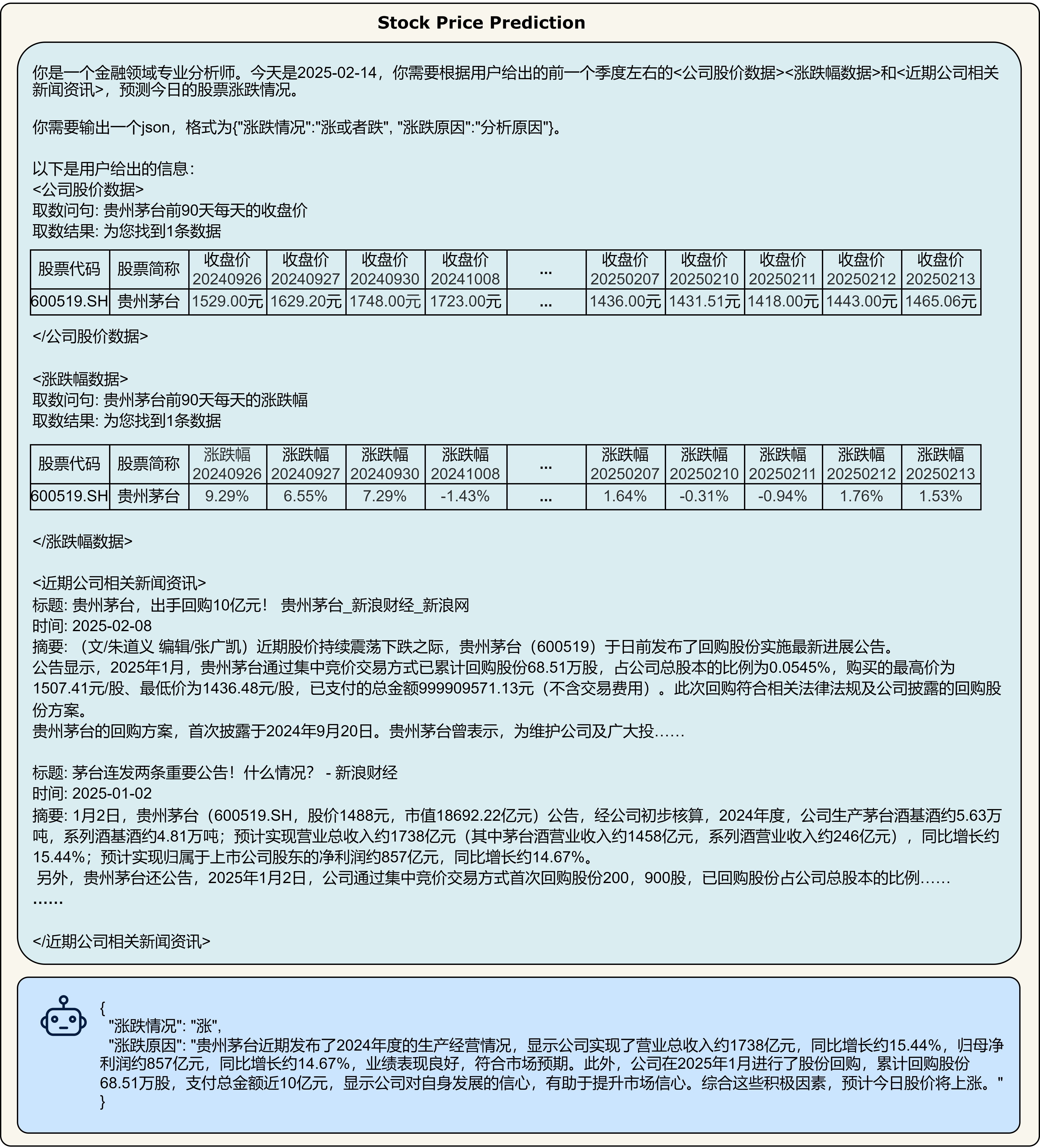}
    \caption{An example instance from the Stock Price Prediction.}
    \label{fig:Stock Price Prediction}
\end{figure*}


\section{Other}
\label{other}

\begin{figure}
    \centering
    \includegraphics[scale=0.090]{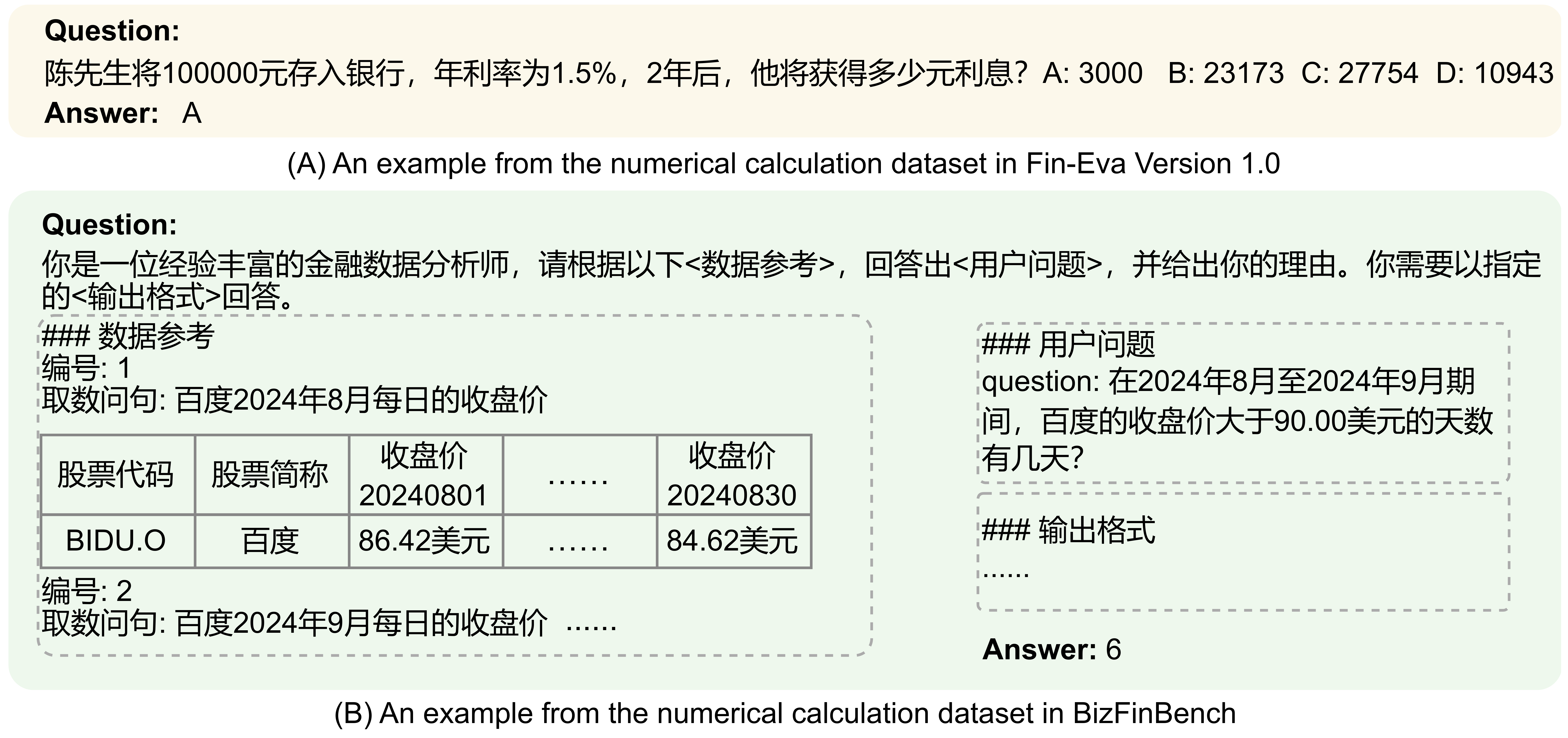}
    \caption{An Chinese Version of the Comparison of Numerical Calculation Questions in Fin-Eva~\cite{team2023FinEva} and \TheName{}}
    \label{fig:comparison}
\end{figure}

\end{document}